\documentclass[sigconf]{acmart}

\usepackage{array}
\usepackage{caption}
\usepackage{enumitem}
\usepackage{booktabs}
\usepackage{multirow} 
\usepackage[caption=false,font=footnotesize,labelfont=sf,textfont=sf]{subfig}

\newcommand{\eat}[1]{}
\usepackage{enumitem}
\setlist[itemize]{leftmargin=*}
\AtBeginDocument{%
  \providecommand\BibTeX{{%
    \normalfont B\kern-0.5em{\scshape i\kern-0.25em b}\kern-0.8em\TeX}}}

\setcopyright{acmcopyright}
\copyrightyear{2024}
\acmYear{2024}
\acmDOI{XXXXXXX.XXXXXXX}

\acmConference[WWW ’24]{Proceedings of the ACM Web Conference 2024}{May 13--17,
  2024}{Singapore, Singapore}
\acmBooktitle{Proceedings of the ACM Web Conference 2024 (WWW'24), May 13--17, 2024, Singapore, Singapore} 
\acmPrice{15.00}
\acmISBN{978-1-4503-XXXX-X/18/06}

\begin{document}

% \title{Knowledge-Enhanced Anomaly Detection via Cross-Space Alignment}

% \title{Weakly Supervised Anomaly Detection via\\ Knowledge-Data Alignment}
\title{Weakly Supervised Anomaly Detection via\\ Knowledge-Data Alignment}

\author{Haihong Zhao}
\affiliation{%
  \institution{Hong Kong University of Science and Technology (Guangzhou)}
  \streetaddress{}
  \city{Guangzhou}
  \country{China}
  }
\email{hzhaobf@connect.ust.hk}

\author{Chenyi Zi}
\affiliation{%
  \institution{Hong Kong University of Science and Technology (Guangzhou)}
  \streetaddress{}
  \city{Guangzhou}
  \country{China}
  }
\email{czi447@connect.hkust-gz.edu.cn}

\author{Yang Liu}
\affiliation{%
  \institution{Hong Kong University of Science and Technology}
  \streetaddress{}
  \city{Hong Kong}
  \country{China}
  }
\email{yliukj@connect.ust.hk}

\author{Chen Zhang}
\affiliation{%
  \institution{CreateLink Technology}
  \streetaddress{}
  \city{}
  \country{China}
  }
\email{zhangchen@chuanglintech.com}

\author{Yan Zhou}
\affiliation{%
  \institution{CreateLink Technology}
  \streetaddress{}
  \city{}
  \country{China}
  }
\email{zhouyan@chuanglintech.com}

\author{Jia Li\textsuperscript{\rm *}}\thanks{\rm * Corresponding author}
\affiliation{%
  \institution{Hong Kong University of Science and Technology (Guangzhou)}
  \streetaddress{}
  \city{Guangzhou}
  \country{China}
  }
\email{jialee@ust.hk}

\begin{abstract}

Anomaly detection (AD) plays a pivotal role in numerous web-based applications, including malware detection, anti-money laundering, device failure detection, and network fault analysis. Most methods, which rely on unsupervised learning, are hard to reach satisfactory detection accuracy due to the lack of labels. Weakly Supervised Anomaly Detection (WSAD) has been introduced with a limited number of labeled anomaly samples to enhance model performance. Nevertheless, it is still challenging for models, trained on an inadequate amount of labeled data, to generalize to unseen anomalies.
In this paper, we introduce a novel framework Knowledge-Data Alignment (KDAlign) to integrate rule knowledge, typically summarized by human experts, to supplement the limited labeled data. Specifically, we transpose these rules into the knowledge space and subsequently recast the incorporation of knowledge as the alignment of knowledge and data. To facilitate this alignment, we employ the Optimal Transport (OT) technique. We then incorporate the OT distance as an additional loss term to the original objective function of WSAD methodologies. Comprehensive experimental results on five real-world datasets demonstrate that our proposed KDAlign framework markedly surpasses its state-of-the-art counterparts, achieving superior performance across various anomaly types.
\end{abstract}






\maketitle

% \begin{figure}[t]
% \begin{center}
%     \includegraphics[width=0.45\textwidth]{Figures/Figure_1_Claim_the_Problem_Point_Version_New_Description.pdf}
% \end{center}
% \caption{An example to show the role of Knowledge in WSAD. The green dashed line shows that knowledge can be regarded as independent extrinsic information, which is widely used by existing knowledge-driven WSAD methods. The red dashed line indicates that the unknown potential data imagined by people may help supplement the training data via intrinsic integration, which is overlooked. The feature vector of the right data point could be $\{f_{1},f_{2},...,f_{d}, state\}=\{0.27, 0.30, ..., 0.12, 1\}$, and the left data point could be $\{f_{1},f_{2},...,f_{d}, state\}=\{?, ?, ..., ?, 1\}$.}
% \label{fig:role of knowledge in WSAD}
% \vspace{-5mm}
% \end{figure}

\begin{figure*}[t]
\begin{center}
    \includegraphics[width=0.95\textwidth]{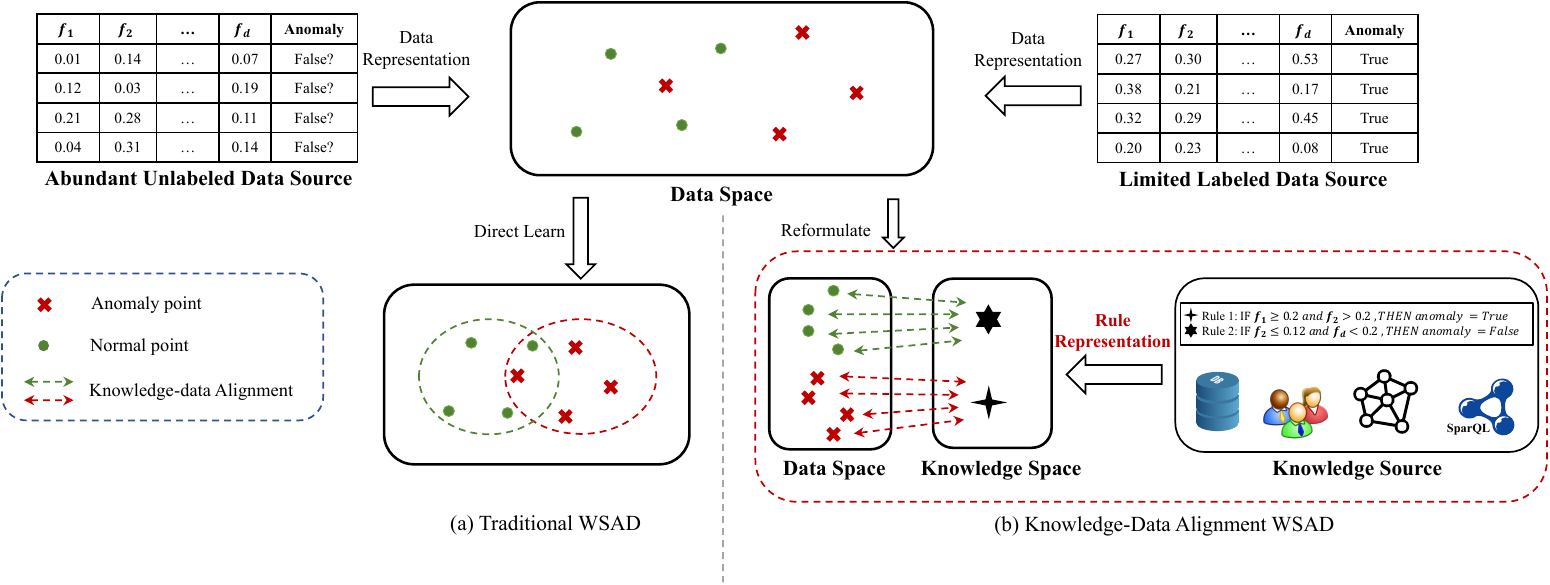}    
\end{center}
\caption{Comparison between traditional WSAD approach (a) and our proposed knowledge-data alignment WSAD framework (KDAlign) (b). We can find that the traditional WSAD mainly focuses on learning from limited labeled data, while our proposed framework introduces knowledge as extra information to supplement limited labeled via knowledge-data alignment. Note that the samples in the unlabeled data source are usually regarded as normal samples, though the unlabeled data may be contaminated by noise~\cite{jiang2023OptimizationParadigm8,pang2021fewshot1}.}
\label{fig:comparison with KDAlign}
% \vspace{-5mm}
\end{figure*}

\section{Introduction}
\label{sec:introduction}

Anomaly detection (AD), aiming at identifying patterns or instances that deviate significantly from the expected behavior or normal patterns, is crucial to extensive web-based applications including malware detection~\cite{khan2019malwaredetection}, anti-money laundering~\cite{lee2020antimoney}, device failure detection~\cite{sipple2020devicefailuredetection}, network fault analysis~\cite{zhao2023RegularizationTermKEGNN,ji2022bidirectional}. Given that labeled anomaly data is typically scarce or costly to acquire, unsupervised methodologies that operate on entirely unlabeled data have gained widespread use. However, in the absence of supervision, these models may incorrectly classify noisy or unrelated data as anomalies, leading to high detection errors.

To alleviate the above issue, Weakly Supervised Anomaly Detection (WSAD) has been proposed to enhance detection accuracy with limited labeled anomaly samples and a large amount of unlabeled data~\cite{jiang2023fewshot2}, shown in Fig.~\ref{fig:comparison with KDAlign}(a). Early studies use unsupervised AD algorithms as feature extractors and learn a supervised classifier with label data~\cite{ruff2019WSADLearningParadigm1, pang2018WSADLearningParadigm2, li2022WSADLearningParadigm3,tian2022WSADLearningParadigm4, huang2020WSADLearningParadigm5,sun2023all}. With the development of deep learning, most recent studies focus on end-to-end frameworks that build on multilayer perceptron, autoencoder, and generative adversarial networks, to directly map input data to anomaly scores~\cite{akcay2019ganomalyOptimizationParadigm6, pang2019WSADOptimizationParadigm3,li2019predicting,zhou2021WSADOptimizationParadigm5}. Nevertheless, models trained on insufficient labeled data fail to generalize to novel anomalies or anomalies not observed during training time~\cite{jiang2023OptimizationParadigm8,jiang2023fewshot2}. Although several works have employed active learning or reinforcement learning to reduce the cost of obtaining anomaly labels, they still require an initial set of labeled data to start the learning process, which can be costly and time-consuming~\cite{li2019semi,zha2020WSADOptimizationParadigm2}.

In this work, we propose to incorporate rule knowledge, which is often derived or summarized by human experts~\cite{zhao2023RegularizationTermKEGNN,yu2023neuralsymbolic5,lin2022lAntiDecisionRule1,li2023AntiHighQualityInaccurateRule1,cheng2023antimoneySimilar1}, similar to label annotation but has been largely overlooked, to help complement the limited labeled data, as shown in Fig.~\ref{fig:comparison with KDAlign}(b). Although rules are high-quality and accessible in practice~\cite{lin2022lAntiDecisionRule1,li2023AntiHighQualityInaccurateRule1}, incorporating them is non-trivial for three reasons: (1) knowledge representation: rules are generally represented by if/else statements~\cite{li2023AntiHighQualityInaccurateRule1,lin2022lAntiDecisionRule1}. In the representation space, rules and data lack a direct correlation~\cite{zhao2023RegularizationTermKEGNN,xie2019embedding}, making them unsuitable for directly training the WSAD models; (2) knowledge-data alignment: intuitively, if two rules are close then their corresponding data samples should be also close~\cite{cheng2023antimoneySimilar1}. For example, in anti-money laundering, a group of fraudsters may possess similar patterns and thus have similar data representations~\cite{lin2022lAntiDecisionRule1,cheng2023antimoneySimilar1,li2023AntiHighQualityInaccurateRule1}. Usually, these fraudsters will be detected by identical or similar if/else rule statements in anti-money laundering systems~\cite{cheng2023antimoneySimilar1}. In this work, we reformulate the knowledge incorporation process as the knowledge-data alignment and supplement the traditional data-only optimizations; (3) noisy knowledge: typically, rules are not always accurate~\cite{cheng2023antimoneySimilar1,li2023AntiHighQualityInaccurateRule1, botao2023pseudo}, thus directly aligning them with data may involve noises and resulting in a performance drop. It is still challenging to ensure the model's performance under noisy rules.
% rules are continuously iterated and updated, and some past rules might be erroneous. It is still challenging to ensure the model performance under noisy rules.

To address the above issues, we propose a novel framework for Weakly Supervised Anomaly Detection via \textbf{K}nowledge-\textbf{D}ata \textbf{Align}ment (\textbf{KDAlign}). KDAlign expects to align knowledge and data to complement the data distribution. For the first challenge, KDAlign employs a knowledge encoder to map the rules into an embedding space, thereby allowing knowledge to correlate with data in the numerical domain. For the second and third challenges, KDAlign leverages the Optimal Transport (OT) technique to align knowledge and data. The primary strength of OT lies in its inherent flexibility. It autonomously determines the optimal transport pairings between knowledge and data, thus establishing an intrinsic connection~\cite{motta2019OTFlexAuto,tang2023robust, tangaclOT,weiqiCIKMAlignment}. Specifically, the OT method intrinsically forms a robust framework, enabling a geometrically faithful comparison of probability distributions and facilitating the information transfer between distinct distributions~\cite{genevay2016OTProbability}. Regarding noisy knowledge, when a sample matches a noisy rule, the distance of that sample to some other closely related rules will be farther, resulting in an increased OT distance penalty. To ensure global optimality, the OT distance between this sample and the noisy rule will be constrained by other correct rules, thereby ensuring the performance of KDAlign.
% 为了解决上述的问题，我们提出了一个新的框架用于小样本异常检测通过知识-数据对齐(KDAlign)。如图2(c)所示，KDAlign期望对齐知识和数据进而对知识和数据进行内在整合。具体来说，KDAlign首先将知识和数据之间建立联系的过程转换为了知识-数据的对齐过程，进而将知识及其对应imagined unknown potential data和known few-shot training data建立联系。然后，KDAlign引入了最优传输技术完成知识-数据的对齐，进而在知识和数据驱动模型之间建立了联系，在数据驱动模型内同时引入了知识分布和数据分布。

To sum up, our contributions are three-fold: 
% 总结来说，我们的贡献包括以下四个方面：
\begin{itemize}

\item To the best of our knowledge, this is the first work to incorporate rule knowledge into WSAD, effectively complementing the limited labeled data.

\item We propose a novel Knowledge-Data Alignment Weakly Supervised anomaly detection framework (KDAlign). 
% This framework addresses the nontrivial issue of incorporating the rule knowledge into WSAD.

%\item A viable method is offered to acquire rule knowledge, mitigating the difficulty of acquiring industrial knowledge.

\item The experimental results on five public WSAD datasets indicate our proposed KDAlign are superior to all the competitors. Furthermore, KDAlign achieves strong performance improvements even with $20\%$ noisy rule knowledge.
    
\end{itemize}

\begin{figure*}[t]
\begin{center}
    \includegraphics[width=0.9\textwidth]{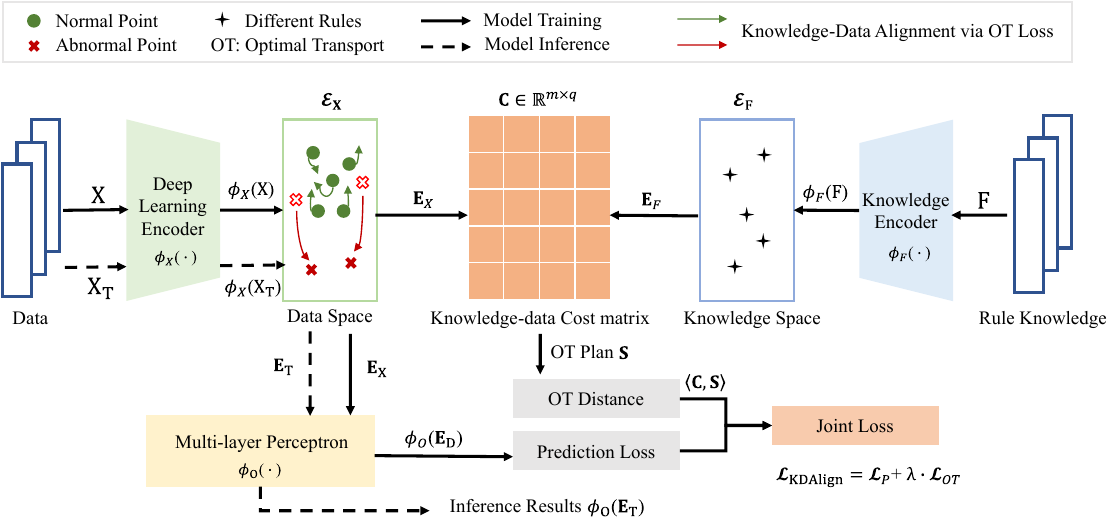}
\end{center}
\caption{Knowledge-data alignment WSAD framework. During the training phase, we firstly use $\phi_{X}$ and $\phi_{F}$ to map $\mathbf{X}$ and $\mathbf{F}$ to two separate embedding spaces and then leverage Optimal Transport (OT) techniques to compute the cost matrix $\mathrm{C}$, thereby obtaining the OT plan $\mathrm{S}$. Next, we compute OT distance \( \langle\boldsymbol{\mathrm{C}}, \boldsymbol{\mathrm{S}}\rangle \) and add it as a loss term to the prediction loss term, forming a joint loss. Finally, we utilize the joint loss to train $\phi_{X}(\cdot)$ and $\phi_{O}(\cdot)$, aligning knowledge and data for incorporating knowledge. In the inference phase of the model, the test data directly yields results by $\phi_{X}$ and $\phi_{O}$. In the data space, both the abnormal and normal points can be aligned via OT.}
\label{fig: overveiw of KDAlign}
\end{figure*}

\section{Preliminary}
\label{sec: preliminary}

\subsection{Rule Knowledge and Logical Formulae}
In this paper, we focus on rule knowledge (if/else). This choice stems from the high-quality and accessible in practice of rules — they present explicit conditions and outcomes. Such transparency allows individuals to understand anomaly and normal data easily. To avoid the potential overlaps among different rules, we adopt a precise knowledge statement format named Logical Formulae. Concretely, logical statements provide a flexible declarative language for expressing structured knowledge (e.g., rule knowledge). In this paper, we focus on \textbf{\textit{propositional logic}}, where a \textbf{\textit{proposition}} $p$ is a statement which is either $\textbf{True}$ or $\textbf{False}$ \cite{klement2004propositional}. A statement (proposition) consists of a subject, predicate and object. It can also be regarded as a ground clause that does not contain any variables \cite{detlovs2011groundclauseslogic}. A \textit{propositional formula} $f$ is a compound of propositions connected by logical connectives \cite{xie2019embedding,CONSOLE2022propositionallogic}, e.g., $\neg$, $\land$, $\lor$, $\Rightarrow$. Also, a propositional formula is equal to a grounding first-order logic formula. In the subsequent content, we use \( \mathbf{F} = \{f_{1}, \ldots, f_{s}\} \) to represent a set of propositional formulae, where \( f_{i} \) is a propositional formula and $s$ is the number of propositional formulae. The concrete proposition formats designed for rule knowledge of AD are introduced in Section \ref{sec:methodology}. 
% \zhao{Define the rule knowledge symbol! add example, is a propositon set.}

\subsection{Problem Statement}
\label{sec:problem statement}

Given a training dataset \( \mathbf{X}=\left\{\mathbf{x}_1, \mathbf{x}_2, \cdots\right. \), \( \left.\mathbf{x}_n, \mathbf{x}_{n+1}, \cdots, \mathbf{x}_{n+m}\right\} \), with \( \mathbf{x}_i \in \mathbb{R}^d \), where \( \mathbf{X_{U}}=\left\{\mathbf{x}_1, \mathbf{x}_2, \cdots, \mathbf{x}_n\right\} \) is a large unlabeled dataset and \( \mathbf{X_{A}}=\left\{\mathbf{x}_{n+1}, \mathbf{x}_{n+2}, \cdots, \mathbf{x}_{n+m}\right\} \) \( (m \ll n) \) is a small set of labeled anomaly examples that often can not cover every possible class of anomaly, a WSAD model \( \mathcal{M} \) is first trained on \( \mathbf{X} \) to output anomaly score $\mathbf{O}:=\mathcal{M}(\mathbf{X}) \in \mathbb{R}^{m \times 1}$, where higher scores indicate a higher likelihood of an abnormal sample. The unlabeled dataset $\mathbf{X_{U}}$ is usually assumed as normal data, though it may be contaminated by some anomalies in practice~~\cite{pang2023WSADOptimizationParadigm4, pang2021fewshot1}. Thus, the trained $\mathcal{M}$ is  required to be robust \textit{w.r.t.} such anomaly contamination. Based on the trained model, we need to predict on the unlabeled test dataset $\textbf{X}_{T} \in \mathbb{R}^{q \times d}$, so to return $\textbf{O}_{T}:=\mathcal{M}(\textbf{X}_{T}) \in \mathbb{R}^{q \times 1}$. In our work, we introduce a set of rule knowledge represented by propositional formulae $\mathbf{F}$, as extra information, to supplement data and then training a WSAD models on \( \{\mathbf{X}, \mathbf{F}\} \).

% previous version
% \textbf{Definition 1} (Weakly Supervised Anomaly Detection) Given  a collection of $(n+m)$ training samples \( \mathbf{X} =\{x_1,...,x_n, x_{n+1}, ..., x_{n+m}\} \in \mathbb{R}^{(n+m) \times d} \) where each sample has $d$ features, and the (binary) ground truth labels of $\mathbf{X}$, i.e., $\mathbf{y}=\{y_1,...,y_n, y_{n+1}, ..., y_{n+m}\} \in \{0,1\}^{n}$, in which $\mathbf{y_{n}}=\{y_1,...,y_n\}$ is the normal label set and $\mathbf{y_{a}}=\{y_{n+1}, ..., y_{n+m}\}$ with $m \ll n$ is a very small anomalous label set, a WSAD model $\mathcal{M}_{WSAD}$ is first trained on $\left\{\mathbf{X}, \mathbf{y}\right\}$ to output anomaly score $\mathbf{O}:=\mathcal{M}_{WSAD}(\mathbf{X}) \in \mathbb{R}^{(n+m) \times 1}$, where higher scores indicate a higher likelihood of an abnormal sample. Note that the normal label set $\mathbf{y_{n}}$ may be contaminated by some anomalies in practice, so trained $\mathcal{M}_{WSAD}$ is often required to be robust \textit{w.r.t.} such anomaly contamination. Based on the trained model, we need to predict on the test samples $\textbf{X}_{test} \in \mathbb{R}^{q \times d}$, so to return $\textbf{O}_{test}:=\mathcal{M}_{WSAD}(\textbf{X}_{test}) \in \mathbb{R}^{q \times 1}$.

\section{Methodology}
\label{sec:methodology}
\subsection{Overview}
Figure~\ref{fig: overveiw of KDAlign} provides an overview of our proposed \textbf{K}nowledge-\textbf{D}ata \textbf{Align}ment WSAD framework (\textbf{KDAlign}). First, we utilize a deep learning encoder and a knowledge encoder to project data $\mathbf{X}$ and knowledge $\mathbf{F}$ into data embedding space $\boldsymbol{\mathcal{E}}_{X}$ and knowledge embedding space $\boldsymbol{\mathcal{E}}_{F}$, respectively, making operations between knowledge and data possible. The deep learning encoder could be based on a multi-layer perceptron, autoencoder, or ResNet-like architecture. The knowledge encoder is a multi-layer graph convolutional network. Second, within the embedding space, we align data and knowledge via the Optimal Transport Technique (OT) and then leverage the alignment result to derive an OT loss term. The OT loss term is subsequently used for introducing the knowledge into deep learning models. Besides, we give an analysis of why KDAlign has the potential to alleviate the noisy knowledge issue. Third, we jointly leverage the OT loss and the prediction loss to train a deep-learning model, expecting to learn better data representations for $\mathbf{X}$ and improve the model performance, where the original loss is computed by the output of the multi-layer perceptron and labels. Then, the trained model can be used for inferring unlabeled test data.
% Figure~\ref{fig: overveiw of KDAlign} provides an overview of our proposed \textbf{K}nowledge-\textbf{D}ata \textbf{Align}ment WSAD framework (\textbf{KDAlign}). First, we utilize deep learning models and a knowledge encoder to project data $\mathbf{X}$ and knowledge $\mathbf{F}$ into data embedding space $\boldsymbol{\mathcal{E}}_{X}$ and knowledge embedding space $\boldsymbol{\mathcal{E}}_{F}$, respectively, making operations between knowledge and data possible. The deep learning model used can be of any structure, such as a multi-layer perceptron, autoencoder, or ResNet-like architecture. The knowledge encoder is a multi-layer graph convolutional network. Second, within the embedding space, we align data and knowledge via the Optimal Transport Technique (OT) and then leverage the alignment result to derive an OT loss term. The OT loss term is subsequently used for introducing the knowledge into deep learning models. Besides, we give an analysis of why KDAlign has the potential to alleviate the noisy knowledge issue. Third, we use the OT loss computed by knowledge-data alignment and prediction loss computed by limited abnormal labels and a large number of normal labels to train a deep-learning model, expecting to learn better data representations for $\mathbf{X}$ and improve the deep learning model performance. Then, the trained model can be used for inferring unlabeled test data.

\subsection{Representation Framework}
\label{sec:knowledge representation}

\subsubsection{\textbf{Data Representation}} Given the training dataset $\mathbf{X}$ with $m$ samples, we use a deep learning encoder to project data into a high-dimensional embedding space, generating the corresponding data embedding set $\mathbf{E}_{X} = \{e_{1}, \ldots, e_{m}\}$, where $m$ is the number of samples. The process is shown by Equation~\ref{eq: data representation}
\begin{equation}
\label{eq: data representation}
\mathbf{E}_{X} = \phi_{X}(\mathbf{X}) \in \mathbb{R}^{m \times h},
\end{equation}
\noindent where $\phi_{X}$ is a deep learning encoder, and $h$ is the dimension of the Data Space $\boldsymbol{\mathcal{E}}_{X}$ defined by \( \mathbf{E}_{X} \). 

% $\phi_{X}$ could be any deep learning model for WSAD, such as ResNet~\cite{gorishniy2021FTTransformerResNet}, DevNet~\cite{zhou2021WSADOptimizationParadigm5}, PReNet~\cite{pang2023WSADOptimizationParadigm4}, and FeaWAD~\cite{zhou2021WSADOptimizationParadigm5}.

\subsubsection{\textbf{Knowledge Representation}}
In order to enable knowledge and data to be operated, we also consider representing knowledge in a high-dimensional embedding space, but embedding knowledge successfully entails rendering it into a format amenable for processing by deep learning methods. Given $s$ if/else statements, we first contemplate transforming their formats into propositional logic and then generate a knowledge set $\mathbf{F}$ with $s$ propositional formulae, which is drawn inspiration from knowledge embedding of logical formulae~\cite{xie2019lensr}.

Next, we provide an example to illustrate how to transform the if/else statement into propositional logic: Given an if/else statement `$If \ attr_1 \ > \ 5 \ and \ attr_2 \ = \ 0 \ , \ then \ anomaly \ is \ True$', where $attr_1$ and $attr_2$ correspond to two attributes of the sample, three propositions $\{p_1=(attr_1 > 5), p_2=(attr_2 = 0), p_3 = (anomaly \ is \ True)\}$ and a proposition formula $f=\{p_1 \land p_2 \Rightarrow p_3\}$ can be generated. The subject, object, and predicate constituting the three propositions are, respectively, $subject=\{attr_1, attr_2, anomaly\}$, $object=\{5, 0, True\}$, $predicate=\{>, =, is\}$.

Building upon the example above, we provide definitions for the subject, object, and predicate in propositional formulae used to describe if/else statements: The subject comprises attribute names present in the sample and the word `anomaly'; the object encompasses attribute thresholds, words `True', and `False'; the predicate consists of numeric relational symbols and the word `is'.

Based on the transformation strategy, we convert if/else statements into a set of propositional formulae \( \mathbf{F} = \{f_1, ..., f_{s}\} \). Subsequently, each propositional formula in \(\mathbf{F}\) is transformed into a graph structure, and then a multi-layer graph convolutional network~\cite{xie2019lensr} is constructed as a knowledge encoder to project propositional formulae into a high-dimensional embedding space named Knowledge Space and generate the knowledge embedding set $\mathbf{E}_{F}=\{e_{1},...,e_{s}\}$, where $s$ is the number of propositional formulae, shown in Equation~\ref{eq:know encoder}
\begin{equation}
\label{eq:know encoder}
\mathbf{E}_{F} = \phi_{F}(\mathbf{F}) \in \mathbb{R}^{s \times h} ,
\end{equation}

\noindent where \(\phi_{F}(\cdot)\) is a multi-layer graph convolutional network, and $h$ is the dimension of the Knowledge Space $\boldsymbol{\mathcal{E}}_{F}$ defined by $\mathbf{E}_{F}$. Since the dimensions of $\boldsymbol{\mathcal{E}}_{F}$ and $\boldsymbol{\mathbf{E}}_{X}$ are both $h$, knowledge and data lie in the same dimensional space. Additionally, $\phi_{F}(\cdot)$ is trained before the training of the deep learning model $\phi_{X}$. The details of the knowledge encoder are shown in the appendix.\\

\subsection{Knowledge-Data Alignment}
\label{sec:knowledge-data alignment for WSAD}

% \subsubsection{Why alignment?}
% \label{sec:why alignment}

% Based on the assumption that if two samples are proximate in the data space, they should similarly be close in the knowledge space, we introduce the knowledge-data alignment, and present the formal definition as follows:\zhao{Pending to update logic...}
\subsubsection{\textbf{Definition}} Given a knowledge set $\mathbf{F}$ and a limited labeled training dataset $\mathbf{X}$, without any observed knowledge-data correspondences, the alignment algorithm returns aligned knowledge-data pairs $M = \{(f_{i}, x_{j}) | (f_{i}, x_{j}) \in F \times X\}$.

\subsubsection{\textbf{Optimal Transport}}
\label{sec:ot for alignment}

To resolve knowledge-data alignment, we leverage the OT technique that has been widely applied in various domains for alignment, such as in image and graph domains~\cite{wang2022OT1,becigneul2020OTApplication2,tang2023robust,zhang2023totAAAI23}. We follow the Kantorovich formulation~\cite{ge2021OTDefinition2,zeng2023ot2}, which can be formally defined in terms of two distributions and a cost matrix as follows

% 简介版本 来源于WWW2023的PARROT论文
% \textbf{Definition 2} (OT and Sinkhorn Distance~\cite{cuturi2013SinkhornDistanceRaw}). Given two discrete distributions $\boldsymbol{\mu}, \boldsymbol{v}$ defined on probability simplex $\Delta_1, \Delta_2$ and a cost matrix $\mathrm{C} \in \mathbb{R}^{n_1 \times n_2}$ measuring the distance between all pairs $\left(x_i, y_j\right) \in \Delta_1 \times \Delta_2$ across two distributions. The OT problem seeks for an optimal coupling/transport plan $\mathrm{S} \in \Pi(\boldsymbol{\mu}, \boldsymbol{v})$ between $\boldsymbol{\mu}$ and $v$ that minimizes the expected cost over the coupling as follows

% 注意观测集合到分布μ的定义，是一个归一化的过程。比如观测集合={1,1,2,3}，对应的μ={1/7, 1/7, 2/7, 3/7}进而可以定义一个4维的概率单纯形。这里的分布μ是一个代理分布，概率单纯形和分布μ一一对应，是将OT问题转为一个convex problem的工具。

\textbf{Definition 3} (Optimal Transport). Given two sets of observations $\boldsymbol{O_1}=\{o_1,...,o_{n_{1}}\}, \boldsymbol{O_2}=\{o_1,...,o_{n_{2}}\}$, there are two discrete distributions $\boldsymbol{\mu}, \boldsymbol{v}$ defined on probability simplex $\Delta_1, \Delta_2$, where $n_1$ is the number of $\boldsymbol{O_1}$, and $n_2$ is the number of $\boldsymbol{O_2}$. Then, a cost matrix $\boldsymbol{\mathrm{C}} \in \mathbb{R}^{n_1 \times n_2}$ is computed for measuring the distance between all pairs $\left(f_i, x_j\right) \in \Delta_1 \times \Delta_2$ across two distributions. The OT problem aims to find an OT plan $\boldsymbol{\mathrm{S}} \in \Pi(\boldsymbol{\mu}, \boldsymbol{v})$ between $\boldsymbol{\mu}$ and $\boldsymbol{v}$ that minimizes the expected cost over the coupling as follows:

\begin{equation}
\label{eq:ot plan computation}
\begin{array}{ll}
\underset{\boldsymbol{\mathrm{S}} \in \Pi(\boldsymbol{\mu}, \boldsymbol{v})}{\min } & \underset{f_i, x_j}\sum \boldsymbol{\mathrm{C}}\left(f_i, x_j\right) \boldsymbol{\mathrm{S}}\left(f_i, x_j\right)=\underset{\boldsymbol{\mathrm{S}} \in \Pi(\boldsymbol{\mu}, \boldsymbol{v})}{\min }\langle\boldsymbol{\mathrm{C}}, \boldsymbol{\mathrm{S}}\rangle, \\ \\
\text { s.t. } & \boldsymbol{\mathrm{S}}\left(f_i,x_j\right) \geq 0, \ for \ all \ i \ and \ j, \\ \\
& \sum_{i=1}^{n_1} \boldsymbol{\mathrm{S}}(f_i,x_j)= \mu_{i}, \sum_{j=1}^{n_2} \boldsymbol{\mathrm{S}}(f_i,x_j)= v_{j},\\
\end{array}
\end{equation}

\noindent where $\boldsymbol{\mathrm{S}}$ is the OT plan, $\Pi(\boldsymbol{\mu}, \boldsymbol{v})$ is the probabilistic coupling between $\boldsymbol{\mu}$ and $\boldsymbol{v}$ (i.e., all the available transport plan between $\boldsymbol{\mu}$ and $\boldsymbol{v}$), $\langle\mathrm{\cdot}, \mathrm{\cdot}\rangle$ is inner product, and corresponding $\langle\boldsymbol{\mathrm{C}}, \boldsymbol{\mathrm{S}}\rangle$ is the Wasserstein distance between $\boldsymbol{\mu}$ and $\boldsymbol{v}$. In this paper, to efficiently solve the OT problem, we employ the Sinkhorn-Knopp algorithm~\cite{cuturi2013SinkhornDistanceRaw,ge2021OTDefinition2}. The objective of the Sinkhorn-Knopp algorithm is to approximate the computation of the Wasserstein distance, enabling efficient computation of Wasserstein distance, particularly in high-dimensional or large-scale scenarios.

In knowledge-data alignment, we regard $\mu$ as the knowledge distribution defined by $\mathbf{F}$ and $v$ as the data distribution defined by $\mathbf{X}$, and then $\boldsymbol{\mathrm{S}}(f_i,x_j)$ indicates the matching score between $f_i$ in the knowledge set $\boldsymbol{F}$ and $x_j$ in $\boldsymbol{X}$. The alignment $M$ can be derived from $\boldsymbol{\mathrm{S}}$:

\begin{equation}
\label{eq:ot alignment}
    M=\underset{M \in \mathbb{M}}{\arg \max } \sum_{\left(f_i, x_j\right) \in M} \boldsymbol{\mathrm{S}}(f_i,x_j),    
\end{equation}

\noindent where $\mathbb{M}$ is the set of all legit alignments, $i=1,...,s$, and $j=1,...,m$.

\subsubsection{\textbf{OT Loss For Weakly Supervised Anomaly Detection}}
\label{sec:ot loss for WSAD}

Obtaining the knowledge-alignment results, we further describe how to utilize the alignment results to benefit WSAD. Firstly, we compute the knowledge-data cost matrix $\boldsymbol{\mathrm{C}} \in \mathbb{R}^{s \times m}$ for measuring the distance between all pairs $\left(\phi_{F}(f_i), \phi_{X}(x_j)\right) \in \boldsymbol{\mathcal{E}}_{F} \times \boldsymbol{\mathcal{E}}_{X}$ across two distribution spaces, which describes the discrepancy between the data embeddings and the knowledge embeddings. Secondly, we can compute the OT plan $S$ and the knowledge-data alignment $M$ by Equations~\ref{eq:ot plan computation} and~\ref{eq:ot alignment}, respectively. Thirdly, we also compute the OT distance, which quantifies the minimum cost required to transform one probability distribution into another, by $\langle\boldsymbol{\mathrm{C}}, \boldsymbol{\mathrm{S}}\rangle$. Finally, the OT distance is used for training the deep learning model.

\paragraph{\textbf{Analysis On Noisy Knowledge Alleviation}}
\label{sec: noisy knowledge alleviation}
Firstly, we need to clarify the effects of noisy rules on WSAD. Since rules and data are matched one-to-one, when noisy rules emerge, it directly leads to detection errors. We would naturally assume that introducing noisy rule information into data embeddings can also impact the performance of the WSAD model. However, benefiting from the OT technique, which takes a global perspective to align rules and data, our proposed KDAlign framework would not be obviously influenced by noisy rules. This is because introducing too much noisy rule information can lead to excessive transport distances between the data and other relevant rules, resulting in a suboptimal transport plan. Therefore, to provide the optimal transport plan, the incorporation of noisy rule knowledge is constrained by other correct rules, thereby alleviating the impact of noisy knowledge.

\subsection{Model Training and Inference}
% \paragraph{Model implementation}

\paragraph{\textbf{Model Training}}
% As mentioned earlier, simply obtaining the results of knowledge-data alignment does not directly enhance deep learning-based AD methods, as the alignment outcomes do not alter the parameters of the deep learning model. The parameters of a deep learning model directly influence data embeddings, i.e., the generation of the Data Space, which in turn affects the classification results of the downstream binary linear classifier $Classifier(\cdot)$. To make the alignment results affect the deep learning model and thereby effectively enhance data embeddings, two feasible approaches exist. One is to add OT distance as a regularization term to the original loss function $\mathcal{L}_{P}$ of the deep learning model. The other is to add OT distance as a loss, $\mathcal{L}_{OT}$, to $\mathcal{L}_{P}$. Since the OT distance calculated through the Sinkhorn-Knopp algorithm is differentiable, we opt for the second approach (i.e., introducing $\mathcal{L}_{OT}$). This addition introduces an auxiliary objective that enables the deep learning method to not only learn data representation based on prediction loss $\mathcal{L}_{P}$ but also minimize $\mathcal{L}_{OT}$, thereby narrowing the gap between data and knowledge representations. 

In addition to deep learning encoders for embedding data, we also employ a multi-layer perceptron (MLP) $\phi_{O}(\cdot)$ to output anomaly scores or classification results based on data embeddings. During the training process, to effectively leverage alignment results, we introduce the OT distance between knowledge and data as a loss term \( \mathcal{L}_{OT} \) added to the prediction loss function \( \mathcal{L}_{P} \) computed by output and sample labels, rather than as a regularization term. This is mainly because the OT distance calculated by the Sinkhorn-Knopp algorithm is differentiable. Concretely, this addition introduces an auxiliary objective that allows both the deep learning encoder and MLP to simultaneously update parameters based on \( \mathcal{L}_{P} \) and \( \mathcal{L}_{OT} \), effectively incorporating knowledge information into data embeddings. The joint loss function $\mathcal{L}_{KDAlign}$ is shown by Equation~\ref{eq: loss function with ot}

\begin{align}
\label{eq: loss function with ot}
    \begin{aligned}
    \mathcal{L}_{KDAlign} = \mathcal{L}_{P} + \lambda \cdot \mathcal{L}_{OT},\\
    \end{aligned}
\end{align}

\noindent where $\mathcal{L}_{P}$ is the prediction loss, and  $\mathcal{L}_{OT}$ is computed by $\langle\boldsymbol{\mathrm{C}}, \boldsymbol{\mathrm{S}}\rangle$, and $\lambda$ is the trade-off factor of $\mathcal{L}_{OT}$. From another perspective, the OT loss offers a targeted optimization direction, thereby effectively incorporating knowledge information and enhancing the model performance. In addition, the prediction loss could also be alternated by other losses for AD (e.g., deviation loss~\cite{pang2019WSADOptimizationParadigm3} and the specially designed deviation loss~\cite{zhou2021WSADOptimizationParadigm5}.

\paragraph{\textbf{Model Inference}}
The trained deep learning encoder $\phi_{X}$ and MLP $\phi_{O}$ comprise the WSAD model $\mathcal{M}$, which is used for inference on test dataset by Equation~\ref{eq: inference process}

\begin{align}
\label{eq: inference process}
    \begin{aligned}
    \mathcal{M}(\mathbf{X_T})=\phi_{O}(\phi_{X}(X_{T})).\\
    \end{aligned}
\end{align}
\noindent where $\phi_{X}$ is the trained deep learning encoder, $\phi_{O}$ is the trained MLP, $\mathcal{M}$ is the trained WSAD model, and $X_{T}$ is the test dataset.

% \paragraph{Model inference}

% 在这个subsection中，我们有意愿去讨论一下我们所提出的框架为什么可以避免噪音数据。首先，我们先明确噪音知识会直接对异常检测任务带来哪些影响。知识和数据是一一匹配的，当噪音知识匹配上一个测试数据的的时候，会直接导致异常检测的错误。然而，我们提出的KDAlign方法是通过对齐的方式去引入知识分布，进而对小样本数据分布进行的补全。他是一种全局的视角，在对齐知识和数据的过程中，噪音知识并不会对数据表示造成太大的影响，因为他受到其他与数据相近的知识的制约。

\section{Experiments}
\label{evaluation}

In this section, we study the experimental results of our proposed method and baselines to answer three research questions:

\begin{itemize}[leftmargin=1.5em]
    \item \textbf{RQ1.} How effective is the proposed KDAlign framework that incorporates knowledge compared with representative baselines in WSAD?
    \item \textbf{RQ2.} How important is the Knowledge-data Alignment in KDAlign?
    \item \textbf{RQ3.} How does noisy knowledge impact the KDAlign?

\end{itemize}

\subsection{Experimental Setup}

\paragraph{\textbf{Datasets.}} We conduct experiments on five real-world datasets \cite{han2022adbench,tang2023gadbench,jiang2023OptimizationParadigm8}. The \textbf{YelpChi} dataset\cite{yelpchidataset,tang2022icmlGAD} is used for finding anomalous reviews which unjustly promote or demote certain products or businesses on Yelp.com. The \textbf{Amazon} dataset\cite{amazondataset,tang2022icmlGAD} seeks to identify the anomalous users paid to write fake product reviews under the Musical Instrument category on Amazon.com. The \textbf{Cardiotocography} dataset\cite{cardiotocographyRaw} targets to detect the pathologic fetuses according to fetal cardiotocographies. The \textbf{Satellite} dataset \cite{satelliteDataset} is collected for distinguishing anomalous satellite images according to multi-spectral values of pixels in 3x3 neighbourhoods. The \textbf{SpamBase} dataset\cite{spambaseDataset} is leveraged to decide spam e-mails on e-mail systems. The descriptions of the five datasets are shown in Table~\ref{tab:data description}.
% Besides, we utilize the knowledge acquisition module introduced in Section~\ref{sec:Knowledge Acquisition} to process the five datasets and acquire specific anomalous rules that can directly identify anomalous samples. Note that in our experiments, we actually used the datasets preprocessed by rules, which will be introduced later.
% \begin{itemize}
%     \item \textbf{YelpChi}~\cite{yelpchidataset}:
%     \item \textbf{Amazon}~\cite{amazondataset}:
%     \item \textbf{Cardiotocography}~\cite{cardiotocographyRaw}:
%     \item \textbf{Satellite}~\cite{satelliteDataset}
%     \item \textbf{SpamBase}~\cite{spambaseDataset}
% \end{itemize}

\paragraph{\textbf{Metrics.}} We choose two widely used metrics to evaluate the performance of all the methods\cite{tang2023gadbench,ding2021fewshot1WWW,han2022adbench}, namely \textbf{AUPRC (Area Under the Precision-Recall Curve)}, and \textbf{Rec@K (Recall at k)}. AUPRC is the area beneath the Precision-Recall curve at different thresholds. \textbf{AUPRC} can be calculated by the weighted mean of precisions at each threshold, where the increase in recall from the previous threshold serves as the weight. \textbf{Rec@K} is determined by calculating the recall of the true anomalies among the top-k predictions that the model ranks with the highest confidence. We set the value of k as the number of actual outliers in the test dataset. It is noteworthy that in this specific scenario, Rec @K is equivalent to both precision at k and the F1 score at k (\textbf{F1@K}).

\begin{table}[t]
\centering
\renewcommand\arraystretch{1.2}
\setlength{\tabcolsep}{1mm}
\caption{Data description of five datasets used in our experiments. Rule-Detect denotes the number of samples that match rules. Rate is computed by \#Rule-Detect/\#Label. }
% \begin{tabular}{m{6.5em}<{\raggedleft} | m{2.0em}<{\centering} m{1.5em}<{\centering} m{2.00em} <{\centering} m{1.5em}<{\centering} m{2.0em}<{\centering} m{3.50em}<{\centering}}
\resizebox{0.45\textwidth}{!}{
\begin{tabular}{ccccccc}
\cmidrule[1.2pt]{1-7}
\textbf{Name}            & \textbf{Size}       & \textbf{\#Feature}        & \textbf{\#Rule}     &\textbf{\#Label}    & \textbf{\#Rule-Detect}       & \textbf{Rate(\%)}   \\
\cmidrule{1-7}
Amazon          & 11944   & 25       & 20        & 821        & 431           & 52.0        \\
Cardiotocography & 2114    & 21       & 14        & 466        & 281           & 60.0        \\
Satellite       & 6435    & 36       & 23       & 2036        & 1015          & 50.0        \\
SpamBase        & 4207    & 57       & 21       & 1679        & 968           & 58.0        \\
YelpChi         & 45954   & 32       & 88       & 6678        & 1475          & 22.0        \\     
\cmidrule[1.2pt]{1-7}
\end{tabular}
}
\label{tab:data description}
\end{table}

\begin{table*}[t]
\centering
\caption{Performance comparison between representative baselines and KDAlign w.r.t. AUPRC and F1@K. The best results are in bold.}
\vspace{-2mm}
\begin{tabular}{m{8.00em}<{\centering} m{3.50em}<{\centering} m{3.50em}<{\centering} m{3.50em}<{\centering} m{3.50em}<{\centering} m{3.50em}<{\centering} m{3.50em}<{\centering} m{3.50em}<{\centering} m{3.50em}<{\centering} m{3.50em}<{\centering} m{3.50em}<{\centering}}
% \toprule
\cmidrule[1.2pt]{1-11}
\multirow{2}{*}{Model} & \multicolumn{2}{c}{Amazon} & \multicolumn{2}{c}{Cardio} & \multicolumn{2}{c}{Satellite}& \multicolumn{2}{c}{SpamBase} & \multicolumn{2}{c}{YelpChi}   \\
\cmidrule(lr){2-3} \cmidrule(lr){4-5} \cmidrule(lr){6-7} \cmidrule(lr){8-9} \cmidrule(lr){10-11}
&PRC  & F1@K & PRC  & F1@K  & PRC & F1@K  & PRC & F1@K& PRC & F1@K\\
\midrule
KNN                 &0.074 &0.071 &0.370 &0.333 &0.326 &0.359 &0.419 &0.406 &0.145 &0.151\\
SVM                 &0.127 &0.013 &0.654 &0.570 &0.312 &0.296 &0.357 &0.283 &0.130 &0.114 \\
DT                  &0.078 &0.065 &0.261 &0.226 &0.320 &0.357 &0.455 &0.431 &0.145 &0.149     \\ \midrule
DeepSAD             &0.137 &0.206 &0.253 &0.312 &0.604 &0.509 &\textbf{0.762} &\textbf{0.686} &0.187 &0.215 \\
KDAlign-DeepSAD     &\textbf{0.201} &\textbf{0.252} &\textbf{0.420} &\textbf{0.462} &\textbf{0.617} &\textbf{0.535} &0.731 &0.637 &\textbf{0.207} &\textbf{0.248} \\ \midrule
REPEN               &0.116 &0.039 &0.452 &0.473 &\textbf{0.726} &0.648 &0.605 &0.589 &\textbf{0.245} &\textbf{0.283} \\
KDAlign-REPEN       &\textbf{0.289} &\textbf{0.290} &\textbf{0.631} &\textbf{0.559} &\textbf{}0.720 &\textbf{0.658} &\textbf{0.608} &\textbf{0.606} &0.181 &0.204\\ \midrule
DevNet              &0.250 &0.316 &0.266 &0.258 &0.647 &\textbf{0.543} &0.416 &0.420 &0.186 &0.211   \\
KDAlign-DevNet      &\textbf{0.487} &\textbf{0.626} &\textbf{0.490} &\textbf{0.516} &\textbf{0.676} &0.533 &\textbf{0.517} &\textbf{0.526} &\textbf{0.195} &\textbf{0.218}   \\ \midrule
PReNet              &0.580 &0.574 &0.602 &0.591 &0.331 &0.303 &0.848 &0.783 &0.174 &0.200 \\
KDAlign-PReNet      &\textbf{0.728} &\textbf{0.716} &\textbf{0.671} &\textbf{0.624} &\textbf{0.677} &\textbf{0.604} &\textbf{0.850} &\textbf{0.783} &\textbf{0.181} &\textbf{0.205}   \\ \midrule
FeaWAD              &0.779 &0.768 &0.622 &0.591 &0.322 &0.418 &0.749 &0.620 &0.184 &0.220 \\
KDAlign-FeaWAD      &\textbf{0.789} &\textbf{0.794} &\textbf{0.664} &\textbf{0.624} &\textbf{0.601} &\textbf{0.555} &\textbf{0.776} &\textbf{0.734} &\textbf{0.216} &\textbf{0.247} \\ \midrule
ResNet              &0.770 &0.729 &0.566 &0.612 &0.352 &0.384 &0.756 &0.706 &0.183 &0.203\\
KDAlign-ResNet      &\textbf{0.848} &\textbf{0.768} &\textbf{0.659} &\textbf{0.656} &\textbf{0.604} &\textbf{0.594} &\textbf{0.770} &\textbf{0.734} &\textbf{0.209} &\textbf{0.262} \\
\cmidrule[1.2pt]{1-11}
% \bottomrule
\end{tabular}
\label{tab:answer research question 1}
% \vspace{-2mm}
\end{table*}

\paragraph{\textbf{Baselines.}}
% We compare KDAlign with two groups of WSAD baselines. The first group is three typical AD methods, consisting of (\textbf{k-Nearest Neighbors (KNN)}~\cite{cover1967KNN}, \textbf{Support Vector Machine (SVM)}~\cite{cortes1995SVM}, \textbf{Decision Tree (DT)}~\cite{breiman1984DecisionTreeModel}. The second group are representative WSAD methods, including \textbf{DeepSAD}~\cite{ruff2019WSADLearningParadigm1}, \textbf{REPEN}~\cite{pang2018WSADLearningParadigm2}, \textbf{DevNet}~\cite{pang2019WSADOptimizationParadigm3}, \textbf{PReNet}~\cite{pang2023WSADOptimizationParadigm4}, \textbf{FeaWAD}~\cite{zhou2021WSADOptimizationParadigm5}, \textbf{ResNet}~\cite{gorishniy2021FTTransformerResNet}.
We compare the proposed method with the following baselines and give brief descriptions. The first three are typical AD methods. The rest of them are representative of WSAD methods.
\begin{itemize}
    \item \textbf{k-Nearest Neighbors (KNN)}~\cite{cover1967KNN}. A classification method based on the k nearest neighbors in the training set.
    \item \textbf{Support Vector Machine (SVM)}~\cite{cortes1995SVM}. A classification method based on maximum margin.
    \item \textbf{Decision Tree (DT)}~\cite{breiman1984DecisionTreeModel}. A classification method based on tree structure, and every decision path define an if/else statement.
    \item \textbf{DeepSAD}~\cite{ruff2019WSADLearningParadigm1}. A deep semi-supervised one-class method that enhances the unsupervised DeepSVDD.
    \item \textbf{REPEN}~\cite{pang2018WSADLearningParadigm2}. A neural network based model that utilized transformed low-dimensional representation for random distance based detectors.
    \item \textbf{DevNet}~\cite{pang2019WSADOptimizationParadigm3}. A neural network based model trained by deviation loss.
    \item \textbf{PReNet}~\cite{pang2023WSADOptimizationParadigm4}. A neural network based model that defines a two-stream ordinal regression to learn the relation of instance pairs.
    \item \textbf{FeaWAD}~\cite{zhou2021WSADOptimizationParadigm5}. A neural network based model that incorporates the network architecture of DAGMM~\cite{zong2018DAGMM} with the deviation loss of DevNet.
    \item \textbf{ResNet}~\cite{gorishniy2021FTTransformerResNet}. ResNet-like architecture turns out to be a strong baseline~\cite{jiang2023OptimizationParadigm8}.
\end{itemize}

% 首先是获取知识。需要注意的是这几份数据集没有提供规则，并且由于工业安全和隐私等问题，直接获取定义良好的规则是困难的。因此，我们需要对这几份数据集的规则进行模拟，以获取知识。在我们的实验中，我们对每个数据集都使用additional labels去训练若干个决策树模型，然后抽取出决策树模型的用于判断异常样本的决策路径作为我们的if/else rules。

\paragraph{\textbf{Parameter And Implementation Details.}} Firstly, acquiring knowledge is essential. It is worth noticing that the five datasets do not provide rules, and due to industrial security and privacy issues, obtaining well-defined rules directly is challenging. Therefore, we need to simulate the rules of these datasets to acquire knowledge. In our experiments, we train several decision tree models for each dataset using additional labels, and then extract the decision paths from the decision tree models as our if/else rules. In WSAD, anomaly samples are unbalanced and important, so we focus on the decision paths used for anomaly samples. A more detailed description of knowledge acquisition can be referred to the appendix. Secondly, we divide each dataset into a training set, a validation set, and a test set according to the scale of 7:1:2. To ensure that the rules really provide extra information (e.g., unseen anomaly scenarios) to supplement limited labeled samples, we delete the anomaly samples that match rules from the training set. Besides, for each training set, we only retain 10 labeled anomaly samples, treating the rest of the anomalies and all normal samples as unlabeled data, with the default label being normal samples. Besides, we also consider another three training settings with 1, 3, and 5 labeled anomaly samples.

Our implementation of SVM, KNN, and DT is consistent with the APIs of Sklearn~\cite{scikit-learn}. We keep the default settings of SVM, KNN, and DT given by Sklearn. For representative WSAD methods, DeepSAD, REPEN, DevNet, PReNet, and FeaWAD are consistent with the Benchmark DeepOD~\cite{xu2023deepOD}, and ResNet is implemented based on the design for AD from ~\cite{gorishniy2021FTTransformerResNet}. The default optimizer of each baseline is Adam~\cite{kingma2014adam}. To apply our proposed KDAlign framework, we make slight adjustments to the representative WSAD methods. Concretely, during the forward propagation of these methods, in addition to returning the output of the final layer, they also return the sample hidden representations from the layer before the last one. All the models are tuned to the best performance on the validation set. Our codes are released at \url{https://github.com/KDAlignForWWW2024/KDAlign}.

% \vspace{-5mm}

\begin{table*}[t]
\centering
\caption{Performance comparison between representative baselines and KDAlign under the setting of 1, 3, or 5 labeled anomalies w.r.t AUPRC. `-' indicates that the PReNet model can not handle setting of only 1 labeled anomaly samples.}
% \vspace{-2mm}
\begin{tabular}{m{8.0em}<{\centering} m{1.95em}<{\centering} m{1.95em}<{\centering} m{1.95em}<{\centering} m{1.95em}<{\centering} m{1.95em}<{\centering} m{1.95em}<{\centering} m{1.95em}<{\centering} m{1.95em}<{\centering} m{1.95em}<{\centering} m{1.95em}<{\centering} m{1.95em}<{\centering} m{1.95em}<{\centering} m{1.95em}<{\centering} m{1.95em}<{\centering} m{1.95em}<{\centering}}
% \toprule
\cmidrule[1.2pt]{1-16}
\multirow{2}{*}{Model} & \multicolumn{3}{c}{Amazon} & \multicolumn{3}{c}{Cardio} & \multicolumn{3}{c}{Satellite}& \multicolumn{3}{c}{SpamBase} & \multicolumn{3}{c}{YelpChi}   \\
\cmidrule(lr){2-4} \cmidrule(lr){5-7} \cmidrule(lr){8-10} \cmidrule(lr){11-13} \cmidrule(lr){14-16}
 &1  & 3&5& 1  & 3&5  & 1 & 3&5  & 1 & 3&5& 1 & 3&5\\
\midrule
  KNN    &0.065&0.065&0.080&0.236&0.253  &0.286 &0.323&0.319 &0.319 &0.416&0.416 &0.414 &0.145&0.145 &0.145
  \\
SVM  &0.238  &0.124  &0.093  &0.290  &0.304  &0.293  &0.349  &0.305  &0.313  &0.437  &0.425  &0.536  &0.148  &0.150  &0.165\\
DT &0.065  &0.065  &0.113  &0.229  &0.244  &0.286  &0.318  &0.318  &0.318  &0.426  &0.429  &0.433  &0.145  &0.145  &0.145 
\\ \midrule
   DevNet
&0.128 &0.103 &0.135 &0.267 &0.269 &0.286 &\textbf{0.722} &\textbf{0.717} &\textbf{0.725} &0.416 &0.416 &0.416 &\textbf{0.167} &0.169 &0.169 \\
  KDAlign-DevNet  &\textbf{0.739} &\textbf{0.747} &\textbf{0.570} &\textbf{0.528} &\textbf{0.436} &\textbf{0.550} &0.474 &0.694 &0.686 &\textbf{0.489} &\textbf{0.461} &\textbf{0.494} &0.156 &\textbf{0.200} &\textbf{0.195}\\ \cmidrule(lr){1-16}
 PReNet 
&-&0.152 &0.488 &- &0.390 &0.626 &- &0.344 &0.266 &- &0.689 &\textbf{0.819} &- &0.164 &0.161 \\
  KDAlign-PReNet  
&-&\textbf{0.617} &\textbf{0.623} &- &\textbf{0.656} &\textbf{0.558} &- &\textbf{0.748} &\textbf{0.767} &- &\textbf{0.741} &0.817 &- &\textbf{0.176} &\textbf{0.207}\\ \cmidrule(lr){1-16}
  DeepSAD &0.129 &0.070 &0.104 &0.355 &0.248 &0.259 &\textbf{0.732} &\textbf{0.739} &0.670 &\textbf{0.710} &0.668 &0.602 &0.188 &0.199 &0.197 \\
 KDAlign-DeepSAD &\textbf{0.347} &\textbf{0.514} &\textbf{0.271} &\textbf{0.639} &\textbf{0.480} &\textbf{0.517} &0.717 &0.722 &\textbf{0.735} &0.699 &\textbf{0.778} &\textbf{0.779} &\textbf{0.205} &\textbf{0.208} &\textbf{0.215} \\ \cmidrule(lr){1-16}
 REPEN &0.116 &0.116 &0.116 &0.452 &0.452 &0.452 &0.730 &0.730 &0.730 &0.608 &0.608 &0.608 &\textbf{0.245} &\textbf{0.244} &\textbf{0.245} \\
 KDAlign-REPEN &
\textbf{0.289} & \textbf{0.289} & \textbf{0.289} & \textbf{0.630} & \textbf{0.630} & \textbf{0.630} & \textbf{0.741} & \textbf{0.741} & \textbf{0.741} & \textbf{0.631} & \textbf{0.631} & \textbf{0.631} & 0.181 & 0.180 & 0.209
\\ \midrule
  FeaWAD &0.692 &0.274 &0.734 &0.545 &\textbf{0.527} &\textbf{0.622} &0.582 &0.570 &0.686 &\textbf{0.697} &0.717 &0.756 &0.170 &0.177 &0.212
 \\
 KDAlign-FeaWAD &\textbf{0.738} & \textbf{0.611} & \textbf{0.778} & \textbf{0.654} & 0.489 & 0.551 & \textbf{0.751} & \textbf{0.769} & \textbf{0.529} & 0.615 & \textbf{0.765} & \textbf{0.775} & \textbf{0.212} & \textbf{0.193} & \textbf{0.240}
  \\ \midrule
ResNet
&0.785 &0.629 &0.777 &0.590 &\textbf{0.663} &\textbf{0.700} &0.328 &0.359 &0.353 &\textbf{0.625} &0.601 &0.637 &0.183 &0.175 &0.175
\\
 KDAlign-ResNet &\textbf{0.805} & \textbf{0.744} & \textbf{0.834} & \textbf{0.642} & 0.613 & 0.691 & \textbf{0.696} & \textbf{0.676} & \textbf{0.699} & 0.601 & \textbf{0.621} & \textbf{0.668} & \textbf{0.201} & \textbf{0.207} & \textbf{0.207}
 \\
\cmidrule[1.2pt]{1-16}
% \bottomrule
\end{tabular}
\label{tab:answer research question 1 with different labeled anomalies}
% \vspace{-2.0mm}
\end{table*}

\begin{table*}[t]
\centering
\caption{AUPRC and F1@K results of Ablation Study. KD- represents the WSAD method incorporated knowledge without knowledge-data alignment.}
% \vspace{-2mm}
\begin{tabular}{cccccccccccc}
% \toprule
\cmidrule[1.2pt]{1-12}
\multirow{2}{*}{Labeled Anomalies} & \multirow{2}{*}{Model} & \multicolumn{2}{c}{Amazon} & \multicolumn{2}{c}{Cardio} & \multicolumn{2}{c}{Satellite}& \multicolumn{2}{c}{SpamBase} & \multicolumn{2}{c}{YelpChi}   \\
\cmidrule(lr){3-4} \cmidrule(lr){5-6} \cmidrule(lr){7-8} \cmidrule(lr){9-10} \cmidrule(lr){11-12}
& & PRC  & F1@K & PRC  & F1@K  & PRC & F1@K  & PRC & F1@K& PRC & F1@K\\
\midrule
 \multirow{2}{*}{\textbf{1}} 
  & KD-ResNet     &0.792 &0.793 &0.546 &0.559 &0.581 &0.444 &0.538 &0.588 &0.161 &0.187 \\
  &KDAlign-ResNet &\textbf{0.805} &\textbf{0.800} &\textbf{0.642} &\textbf{0.581} &\textbf{0.696} &\textbf{0.575} &\textbf{0.601} &\textbf{0.643} &\textbf{0.201} &\textbf{0.266 }\\
\midrule
 \multirow{2}{*}{\textbf{3}}
  & KD-ResNet     &0.643 &0.651 &\textbf{0.629} &0.612 &0.560 &0.422 &0.595 &0.565 &0.156 &0.188 \\
  &KDAlign-ResNet &\textbf{0.744} &\textbf{0.742} &0.613 &\textbf{0.645 }&\textbf{0.676} &\textbf{0.592} &\textbf{0.621} &\textbf{0.654} &\textbf{0.207} &\textbf{0.261} \\
\midrule
 \multirow{2}{*}{\textbf{5}}
  & KD-ResNet     &0.812 &0.780 &0.682 &\textbf{0.667} &0.517 &0.410 &0.600 &0.645 &0.168 &0.194 \\
  &KDAlign-ResNet &\textbf{0.834} &\textbf{0.806} &\textbf{0.691} &0.645 &\textbf{0.699} &\textbf{0.577} &\textbf{0.668} &\textbf{0.694} &\textbf{0.207} &\textbf{0.263} \\
\midrule
 \multirow{2}{*}{\textbf{10}}
  & KD-ResNet     &0.715 &0.767 &0.633 &0.612 &0.485 &0.506 &0.706 &0.700 &0.171 &0.191 \\
  &KDAlign-ResNet &\textbf{0.848} &\textbf{0.768} &\textbf{0.659} &\textbf{0.656} &\textbf{0.604} &\textbf{0.59}4 &\textbf{0.770} &\textbf{0.734} &\textbf{0.209} &\textbf{0.262} \\
\midrule
 \multirow{2}{*}{\textbf{Average}}
  & KD-ResNet     &0.741 &0.748 &0.623 &0.613 &0.536 &0.446 &0.610 &0.625 &0.164 &0.190 \\
  &KDAlign-ResNet &\textbf{0.808} &\textbf{0.779} &\textbf{0.651} &\textbf{0.632} &\textbf{0.669} &\textbf{0.585} &\textbf{0.665} &\textbf{0.681} &\textbf{0.206} &\textbf{0.263} \\  
\midrule
 % \multirow{2}{*}{\textbf{Standard Deviation \( \downarrow \)}}
 %  & KD-ResNet     &0.077 &0.065 &0.056 &0.044 &\textbf{0.043} &0.042 &\textbf{0.070} &0.061 &0.007 &0.003 \\
 %  &KDAlign-ResNet &\textbf{0.046} &\textbf{0.029} &\textbf{0.032} &\textbf{0.034} &0.044 &\textbf{0.009} &0.075 &\textbf{0.041} &\textbf{0.003} &\textbf{0.002} \\
\cmidrule[1.2pt]{1-12}
% \bottomrule
\end{tabular}
\label{tab:answer research question 2}
\vspace{-2mm}
\end{table*}

\subsection{Performance Comparison (RQ1)}
Table~\ref{tab:answer research question 1} shows the model performance on 5 datasets w.r.t AUCPR and F1@K. Each dataset contains 10 labeled anomalies. Above all, we verify the effectiveness of the KDAlign framework on various deep learning-based WSAD baselines. The KDAlign based AD methods we proposed generally outperform the corresponding baselines.

Specifically, we have the following observations:
\begin{itemize}
    \item We find that typical anomaly detection methods, including KNN, SVM, and DT, struggle when the number of labeled anomalies is extremely sparse. The SVM method shows good results on the Cardiotocography dataset, which might be coincidental.
    \item We observe that while representative WSAD methods demonstrate impressive performance on certain datasets, they invariably have weak performance on one or more datasets. For instance, DeepSAD outperforms most baselines on the SpamBase dataset but underperforms on the Amazon dataset. Based on the characteristics of these five datasets, the discrepancy may be due to the higher feature count in SpamBase and the lower feature count in the Amazon dataset. This is because DeepSAD focuses on anomaly feature representation learning~\cite{jiang2023fewshot2}. REPEN method outperforms all other methods on the YelpChi dataset, but falls short on both the Amazon and Cardiotocography datasets. We surmise that this is because REPEN is an unsupervised anomaly feature representation learning method~\cite{jiang2023fewshot2}, and datasets like Amazon and Cardiotocography neither offer as many samples as YelpChi nor as many features as Satellite and SpamBase. The performance of the DevNet method on the Amazon, Cardiotocography, and SpamBase datasets is not satisfactory. This is mainly because the labeled anomalies available for these three datasets cover a limited range of anomaly scenarios. As the DevNet approach focuses on Anomaly Score Learning~\cite{jiang2023fewshot2}, the scores it learns fail to distinguish between normal and anomalous samples.
    \item We find that methods like PReNet and FeaWAD, both belonging to the anomaly score learning~\cite{jiang2023fewshot2}, generally outperform DevNet and other baseline methods across all datasets. We attribute this promising performance primarily to the design of the PReNet and FeaWAD. PReNet method takes anomaly-anomaly, anomaly-unlabeled, and unlabeled-unlabeled instance pairs as input, and learns pairwise anomaly scores by discriminating these three types of linear pairwise interactions. This is an augmentation process of data distribution for existing labeled anomalies, which subsequently aids in the learning of the final anomaly scores. The autoencoder architecture of FeaWAD is capable of mapping limited labeled samples to a latent space, thereby extending the distribution of these sparsely labeled anomalies and improving the learned Score distributions.
    \item We observe that our proposed KDAlign framework consistently enhances the performance of representative WSAD methods. For example, the KDAlign-PReNet method exhibits a 104.53\% improvement over PReNet on the Satellite dataset, climbing from the last rank (excluding typical AD methods) to the second rank. In addition, DeepSAD, REPEN and DevNet, which introduced KDAlign, also have nearly doubled improvements on the Amazon and Cardiotocography data sets. Even when FeaWAD and ResNet already demonstrate commendable results, the KDAlign framework still manages to further boost their performance. In some isolated cases, KDAlign fails to enhance the performance of baseline methods. The reason may be that the unseen anomalies learned by the original baseline methods overlap with the anomalies covered by knowledge, and aligning them might distort the original data representation.
    \item We find that the best-performing methods on each dataset are based on the KDAlign framework. This indicates that KDAlign can not only improve the performance of baseline methods but also holds the potential to provide new state-of-the-art results in weakly supervised settings.
\end{itemize}

In addition, we also use Table~\ref{tab:answer research question 1 with different labeled anomalies} to present the experimental results of KDAlign and representative baselines with respect to AUCPR when the number of labeled anomalies is 1, 3, or 5. The results for F1@K will be shown in the appendix.

\subsection{Ablation Study (RQ2)}
We present the results of our Ablation Study in Table~\ref{tab:answer research question 2}, with respect to AUPRC and F1@K. Concretely, we compare the model performance of KDAlign-ResNet and KD-ResNet across five datasets under four WSAD settings, where the KD-ResNet introduces knowledge without the knowledge-data alignment and the label anomalies of each dataset are respectively 1,3,5 and 10. According to the experimental results, we can clearly find that KDAlign-ResNet outperforms KD-ResNet in almost all settings. Besides, we observe that the standard deviation of KDAlign-ResNet across the four settings is significantly lower than that of KD-ResNet. This suggests that the performance of KDAlign-ResNet remains relatively consistent as the number of labeled samples varies from 1 to 10, reflecting the robustness of the KDAlign framework.
% 我们使用Table 3展示了我们的Ablation Study的实验结果，w.r.t AUPRC and F1@K。In detail，我们在五个数据集的4种WSAD设置下对比了KDAlign-ResNet方法和不使用知识-数据对齐的引入知识的KD-ResNet方法的模型性能，其中每个数据集的标签异常分别是1,3,5和10个。根据实验结果，我们可以清楚地发现：(1) 我们发现几乎在所有情况下KDAlign-ResNet方法outperforms KD-ResNet。（2）我们发现KDAlign-ResNet在4中settings下的标准差比KD-ResNet要显著的地，说明KDAlign-ResNet在标签样本从1到10变化的时候，性能没有较大的波动，体现了KDAlign框架的鲁棒性。

\begin{figure}[h]
% \vspace{-2mm}
\centering
\includegraphics[width=0.46\textwidth]{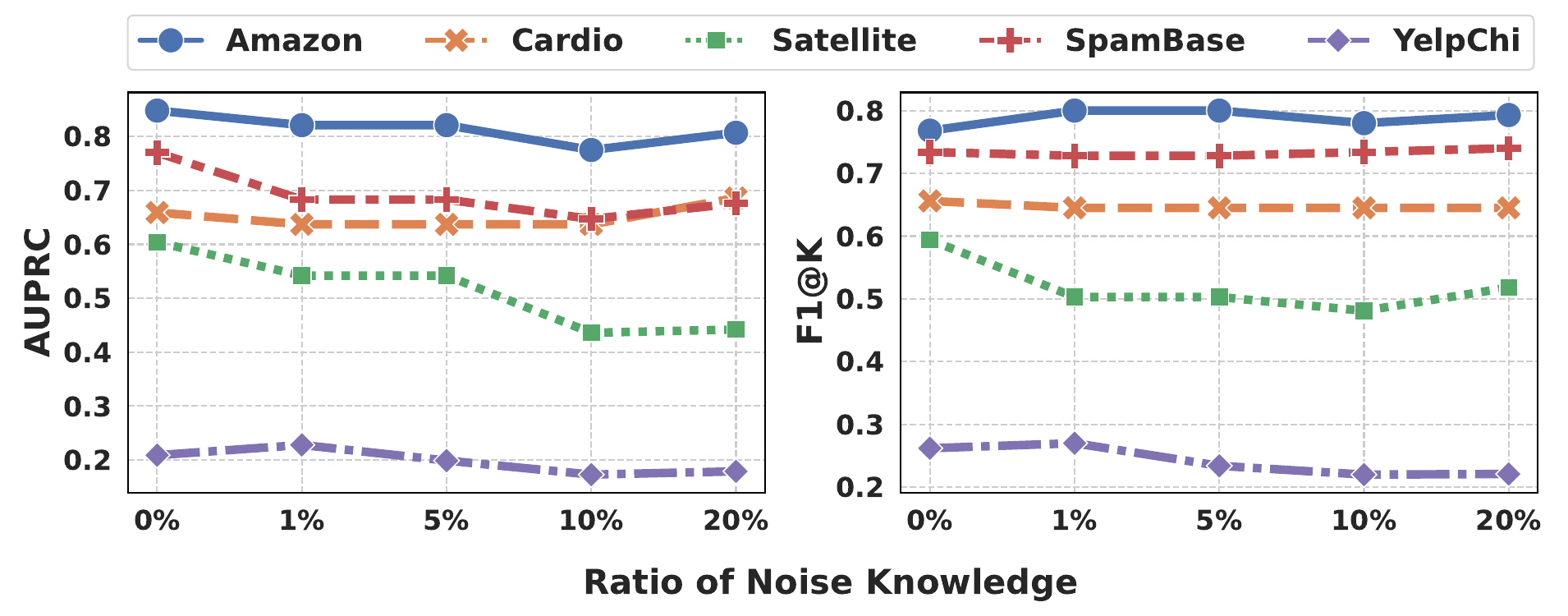}
% \vspace{-1mm}
\caption{Noisy knowledge study on KDAlign-ResNet.}
\label{fig:answer research question 3}
\vspace{-4mm}
\end{figure}

\subsection{Impact of Noisy Knowledge (RQ3)}
We use Figure~\ref{fig:answer research question 3} to illustrate the impact of noisy knowledge on the performance of KDAlign-ResNet. By `noisy knowledge', we refer to the incompletely correct rules that will incorrectly judge some samples, leading to detection errors. Specifically, we investigate the performance of KDAlign-ResNet across five datasets under four noisy settings. The ratio of noise knowledge pertains to the ratio of incompletely correct rules to the total rules. From our experiments, we make the following observations: Compared to the setting without noise, we find that as the ratio increases, the performance of KDAlign-ResNet does not fluctuate significantly, except for the Satellite dataset. The reason might be the knowledge with noise happens to cover some unseen important anomaly scenarios, which in turn results in a decline in model performance. It is worth noting that compared with Table~\ref{tab:answer research question 1}, even when impacted by noisy knowledge, the performance of KDAlign-ResNet remains superior to that of ResNet.
% 我们使用Figure 3展示the impact of noisy knowledge on the performance of KDAlign-ResNet. 噪音知识指的是不完全正确的规则，其会错误的判断一些样本，导致检测的错误。具体而言，我们探究了KDAlign-ResNet在5个数据集上的4种噪音场景下的模型性能。The ratio of noise knowledge 指的是不完全正确的规则的数量。通过实验，我们有如下的观察：对比没有噪音的场景，我们发现随着噪音比率的增加，KDAlign-ResNet的性能并没有显著的波动，除了Satellite数据集。这可能是因为出现噪音的知识刚好覆盖的是一些unseen anomaly scenarios，进而导致了模型性能的衰减。值得注意的是，对比Table~\ref{tab:answer research question 1}，即使是受到了噪音知识的影响，KDAlign-ResNet的性能仍然是由于ResNet的。

\section{Related Work}
\label{sec: related work}
\subsection{Weakly Supervised Anomaly Detection}
Weakly supervised anomaly detection aims to train an effective AD model with limited labeled anomaly samples and extensive unlabeled data. Early studies~\cite{ruff2019WSADLearningParadigm1,ruff2018deepsvdd,pang2018WSADLearningParadigm2} on WSAD primarily involved designing a feature extractor based on unsupervised AD algorithms and then learning a supervised classifier using labeled data such as eepSAD~\cite{ruff2019WSADLearningParadigm1} and REPEN~\cite{pang2018WSADLearningParadigm2}. Recent studies~\cite{li2022dualMGAN,pang2023WSADOptimizationParadigm4,zhou2021WSADOptimizationParadigm5,li2022semi} focus on designing end-to-end deep framework. For example, 
% Dual-MGAN~\cite{li2022dualMGAN} combines several GANs to establish reference distribution and augment data for detecting both discrete and grouped anomalies. 
DevNet~\cite{pang2019WSADOptimizationParadigm3} utilizes a prior probability and a margin hyperparameter to enforce obvious deviations in anomaly scores between normal and abnormal data. 
FeaWAD~\cite{zhou2021WSADOptimizationParadigm5} incorporates the DAGMM~\cite{zong2018DAGMM} network architecture with the deviation loss.
PReNet~\cite{pang2023WSADOptimizationParadigm4} formulates the scoring function as a pairwise relation learning task.

Another research line utilizes active learning or reinforcement learning to reduce the cost of obtaining anomaly labels. 
% they still need some initial labels to start the learning process, which is costly and time-consuming. 
For instance, AAD~\cite{das2016AAD} leverages the active learning technique, which operates in an interactive loop for data exploration and maximizes the total number of true anomalies presented to the expert under a query budget. DPLAN~\cite{pang2021WSADOptimizationParadigm1} considers simultaneously exploring both limited labeled anomaly examples and scarce unlabeled anomalies to extend the learned abnormality, leading to the joint optimization of both objectives. 

% In addition, some supervised methods, such as ResNet~\cite{gorishniy2021FTTransformerResNet}, not specifically designed for AD, can also be adapted for WSAD. However, a potential issue is that ground truth labels may not be entirely accurate (i.e., there often exist some unlabeled anomaly noises in normal samples) to capture anomaly scenarios. Therefore, these supervised methods might fail to detect unseen anomalies.

% In general, existing WSAD models mainly focus on leveraging limited labeled data more effectively for deriving better AD models. 
In contrast to above studies, our work introduces rule knowledge to supplement the limited anomaly samples. Similar to label annotations, such knowledge also contains human supervision, but has been largely overlooked. 

\subsection{Neural-symbolic Systems}
The symbolic system excels in leveraging knowledge, while the neural system is adept at harnessing data. Both knowledge and data play a pivotal role in decision-making processes. There is a burgeoning interest among AI researchers to fuse the symbolic and neural paradigms, aiming to harness the strengths of both~\cite{garcez2022neuralsymbolic1,kaur2018neuralsymbolic2,sourek2018neuralsymbolic3,dragone2021neuralsymbolic4,xie2019lensr}. When juxtaposing neural-symbolic systems against purely neural or symbolic ones, three aspects come to the fore~\cite{yu2023neuralsymbolic5}. First is the Efficiency. Neural-symbolic models can expedite computations, making them suitable for reasoning on vast data sets. Second is the Generalization. These systems are not solely reliant on extensive labeled datasets, endowing them with impressive generalization capabilities. By integrating expert or background knowledge, neural-symbolic models can compensate for sparse training data, achieving commendable performance without sacrificing generalizability. Third is the interpretability. Neural-symbolic architectures offer transparency in their reasoning, enhancing their interpretability~\cite{yu2023neuralsymbolic5}. Such transparency is invaluable in fields like medical image analysis, where stakeholders require both the outcome and an understanding of the decision-making rationale~\cite{yu2023neuralsymbolic5}. In general, the neural-symbolic system is a promising approach to effectively simultaneously leverage knowledge and data for decision-making processes. However, its potential in weakly supervised anomaly detection has yet to be explored.

% \begin{figure}[t]
% % \vspace{-2mm}
% \centering
% \includegraphics[width=0.46\textwidth]{Figures/KDAlign_ResNet_noise_knowledge_analysis.pdf}
% % \vspace{-2mm}
% \caption{Noisy knowledge study on KDAlign-ResNet.}
% \label{fig:answer research question 3}
% % \vspace{-4mm}
% \end{figure}

\section{Conclusion and Future work}
In this paper, we study the problem of weakly supervised anomaly detection and propose a novel WSAD framework named KDAlign, which reformulates knowledge incorporation as knowledge-data alignment, adopts OT for effectively resolving knowledge-data alignment, and finally supplements the limited anomaly samples to improve the performance of WSAD models. We extensively conduct experiments on five real-world datasets and the experimental results demonstrate that our framework outperforms the other competitors.

For the future, we plan to extend our work in following directions: (1) Extend to graph domain~\cite{jiang2023fewshot2}; (2) Introduce other OT methods, such as~\cite{tang2023robust}; (3) Improve explainability.

% In future, we plan to extend our work in following directions: (1) Extending KDAlign to the WSAD tasks with other data modalities (e.g., graph)~\cite{jiang2023fewshot2}. (2) Introducing other OT techniques~\cite{tang2023robust}. (3) Evaluating the performance of training WSAD models and knowledge encoders at the same time. (4) Evaluating KDAlign under different settings (i.e., unsupervised and supervised learning). (5) Giving some explainability analysis.

% In this paper, we study the problem of weakly supervised anomaly detection and propose a novel WSAD framework named KDAlign, which reformulates knowledge incorporation as knowledge-data alignment, adopts OT for effectively resolving knowledge-data alignment, and finally supplements the limited anomaly samples to improve the performance of WSAD models. We extensively conduct experiments on five real-world datasets and the experimental results demonstrate that our framework outperforms the other competitors. In the future, we may consider extending KDAlign to the WSAD tasks with other data modalities (e.g., graph).

% In future, we plan to extend our work in following directions: (1)

% Acknowledgments start
\begin{acks}
This work was supported by NSFC Grant No. 62206067, HKUST(GZ)-Chuanglin Graph Data Joint Lab and Guangzhou-HKUST(GZ) Joint Funding Scheme 2023A03J0673.
\end{acks}
% Acknowledgments end

\bibliographystyle{ACM-Reference-Format}
\bibliography{www-2024}

\clearpage

\appendix

\section{Knowledge Acquisition}
\label{sec:Knowledge Acquisition}

Rule knowledge usually widely exists in industry~\cite{zhao2023RegularizationTermKEGNN}. However, hindered by concerns for industrial safety and privacy, procuring traditional rule knowledge from the industry poses challenges. Therefore, it is necessary to find an alternate way to simulate the industrial rule knowledge. We find the decision tree method is a promising way~\cite{quinlan1987DecisionTreeAsRules,bouzida2006DecisionTreeForAD1}, where every decision path can be regarded as a rule knowledge. First, the representation format of decision paths is the same as industrial rule knowledge, often manifesting as if/else statements. Second, decision paths are conveniently accessible—for instance, we can extract decision paths from well-trained decision trees.

\begin{figure}[ht]
\begin{center}
    \includegraphics[width=0.45\textwidth]{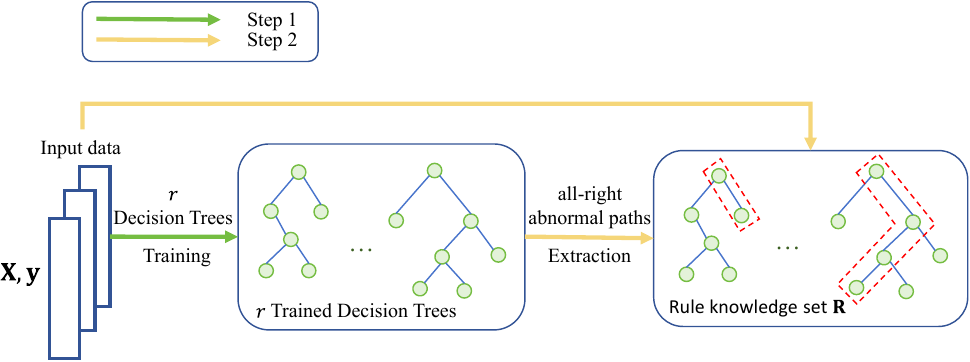}
\end{center}
\caption{Knowledge Acquisition: Using decision trees to simulate industrial rule knowledge.}
\label{fig:knowledge acquisition}    
\vspace{-5mm}
\end{figure}

Specifically, referring to Fig.~\ref{fig:knowledge acquisition}, we assign three steps to acquire rule knowledge based on decision tree models: \textbf{Step 1:} Given a collection of $m$ samples $\mathbf{X} =\{x_1,...,x_{m}\} \in \mathbb{R}^{m \times d}$ and the binary ground truth labels  $\mathbf{y}=\{y_1,...,y_m\} \in \{0,1\}^{m}$, we train $r$ decision trees; \textbf{Step 2:} For the $r$ trained decision trees, we extract \textit{all-right anomaly paths} (knowledge set) $\mathbf{R}=\{r_1,...,r_{s}\}$ included in them as the rule knowledge. An \textit{all-right anomaly path} means that the labels of specific samples in $\mathbf{X}$ passing the decision path are all anomalous.
% All the samples in \textit{all-right abnormal paths} constitute a knowledge-sample set $\mathbf{X_{K}} \in \mathbb{R}^{n_{know} \times d} (n_{know} < n)$ and a ground truth label set $\mathbf{y_{know}} \in \{0,1\}^{n_{know}}$; \textbf{Step 3:} Based on $\mathbf{X}$, we deleted the samples in $\mathbf{X_{know}}$ and generate a new collection of $n_2$ samples $\mathbf{\Hat{X}}$ and a corresponding label set $ \mathbf{\hat{y}}$, which is computed by Equation~\ref{eq: isolate failed matched samples}. In this way, we obtain a rule knowledge set $\mathbf{K}$. The concrete function of \textbf{Step 3} and  will be explained in Section~\ref{sec:knowledge representation}.

% \begin{align}
% \label{eq: isolate failed matched samples}
%     \begin{aligned}
%     \mathbf{\Hat{X}} &= \mathbf{X} - \mathbf{X_{know}}\\
%     \mathbf{\hat{y}} &= \mathbf{y} - \mathbf{y_{know}}\\
%     \end{aligned}
% \end{align}

% \noindent where $\mathbf{\Hat{X}}$ is set difference (`$-$') between the original collection set $\mathbf{X}$ and knowledge-sample set $\mathbf{X_{know}}$, and $\mathbf{\hat{y}}$ is set difference (`$-$') between the ground truth labels $\mathbf{y}$, $\mathbf{y_{know}}$ of $\mathbf{X}$ and $\mathbf{X_{know}}$.

\begin{table*}[t]
\centering
\caption{Performance comparison between representative baselines and KDAlign under the setting of 1, 3, or 5 labeled anomalies w.r.t F1@K. `-' indicates that the PReNet model can not handle the setting of only 1 labeled anomaly sample.}
\vspace{-2mm}
\resizebox{1.0\textwidth}{!}{
\begin{tabular}{m{8.0em}<{\centering} m{1.95em}<{\centering} m{1.95em}<{\centering} m{1.95em}<{\centering} m{1.95em}<{\centering} m{1.95em}<{\centering} m{1.95em}<{\centering} m{1.95em}<{\centering} m{1.95em}<{\centering} m{1.95em}<{\centering} m{1.95em}<{\centering} m{1.95em}<{\centering} m{1.95em}<{\centering} m{1.95em}<{\centering} m{1.95em}<{\centering} m{1.95em}<{\centering}}
% \toprule
\cmidrule[1.2pt]{1-16}
\multirow{2}{*}{Model} & \multicolumn{3}{c}{Amazon} & \multicolumn{3}{c}{Cardio} & \multicolumn{3}{c}{Satellite}& \multicolumn{3}{c}{SpamBase} & \multicolumn{3}{c}{YelpChi}   \\
\cmidrule(lr){2-4} \cmidrule(lr){5-7} \cmidrule(lr){8-10} \cmidrule(lr){11-13} \cmidrule(lr){14-16}
 &1  & 3&5& 1  & 3&5  & 1 & 3&5  & 1 & 3&5& 1 & 3&5\\
\midrule
  KNN    &0.052&0.045&0.071&0.204&0.215&0.226&0.357&0.357&0.357  &0.420 &0.417&0.411 &0.150 &0.149&0.149
  \\
SVM  
&0.045  &0.026  &0.026  &0.269  &0.280  &0.215  &0.308  &0.279  &0.318  &0.431  &0.386  &0.440  &0.162  &0.153  &0.171
\\
DT &0.052  &0.052  &0.097  &0.194  &0.194  &0.226  &0.355  &0.357  &0.355  &0.426  &0.426  &0.420  &0.150  &0.150  &0.150
\\ \midrule
   DevNet
&0.180 &0.116 &0.123 &0.247 &0.269 &0.258 &\textbf{0.555} &0.557 &\textbf{0.562} &0.420 &0.423 &0.423 &\textbf{0.181} &0.185 &0.184
\\
  KDAlign-DevNet  &\textbf{0.735} &\textbf{0.709} &\textbf{0.638} &\textbf{0.528} &\textbf{0.436} &\textbf{0.516} &0.459 &\textbf{0.559} &0.547 &\textbf{0.489} &\textbf{0.461} &\textbf{0.494} &0.156 &\textbf{0.200} &\textbf{0.195}\\ \cmidrule(lr){1-16}
 PReNet &-&0.277 &0.568 &- &0.419 &0.581 &- &0.386 &0.244 &- &0.669 &0.754 &- &0.173 &0.194
 \\
  KDAlign-PReNet  
&-&\textbf{0.658} &\textbf{0.600} &- &\textbf{0.634} &\textbf{0.591} &- &\textbf{0.597} &0.641 &- &0.703 &0.749 &- &0.221 &0.224\\ \cmidrule(lr){1-16}
  DeepSAD &0.110 &0.077 &0.161 &0.387 &0.258 &0.247 &0.582 &0.577 &0.548 &0.651 &0.629 &0.594 &0.209 &0.221 &0.221
\\
 KDAlign-DeepSAD&\textbf{0.355} & \textbf{0.484} & \textbf{0.368} & \textbf{0.624} & \textbf{0.495} & \textbf{0.581} & \textbf{0.641} & \textbf{0.611} & \textbf{0.643} & \textbf{0.706} & \textbf{0.723} & \textbf{0.689} & \textbf{0.230} & \textbf{0.248} & \textbf{0.259}
 \\ \cmidrule(lr){1-16}
 REPEN &0.039 &0.039 &0.039 &0.473 &0.473 &0.473 &0.655 &0.655 &0.655 &0.589 &0.589 &0.589 &\textbf{0.283}&\textbf{0.282} &\textbf{0.283}\\
 KDAlign-REPEN 
&\textbf{0.290} & \textbf{0.290} & \textbf{0.290} & \textbf{0.559} & \textbf{0.559} & \textbf{0.559} & \textbf{0.658} & \textbf{0.658} & \textbf{0.658} & \textbf{0.608} & \textbf{0.608} & \textbf{0.608} & 0.181 & 0.180 & 0.209
\\ 
\midrule
  FeaWAD 
&0.677 &0.368 &0.710   &0.570   &\textbf{0.548}  &\textbf{0.656}  &0.472  &0.457  &\textbf{0.538}  &\textbf{0.677}  &0.703  &0.726  &0.199  &0.204  &0.235\\
 KDAlign-FeaWAD &\textbf{0.729} &\textbf{0.677} &\textbf{0.774}  &\textbf{0.602}  &0.505  &0.581  &\textbf{0.623}  &\textbf{0.650}   &0.501  &0.640   &\textbf{0.754}  &\textbf{0.740}   &\textbf{0.245}  &\textbf{0.228}  &\textbf{0.266} \\
\midrule
ResNet
&0.781 &0.658 &0.774  &\textbf{0.602}  &0.645  &\textbf{0.677}  &0.340   &0.369  &0.369  &0.620   &0.586  &0.603  &0.195  &0.190   &0.207
\\
 KDAlign-ResNet &\textbf{0.800} & \textbf{0.742} & \textbf{0.806} & 0.581 & 0.645 & 0.645 & \textbf{0.575} & \textbf{0.592} & \textbf{0.577} & \textbf{0.643} & \textbf{0.654} & \textbf{0.694} & \textbf{0.266} & \textbf{0.261} & \textbf{0.263}
\\
\cmidrule[1.2pt]{1-16}
% \bottomrule
\end{tabular}
}
\label{tab:answer research question 1 with 5 labeled anomalies}
% \vspace{-5mm}
\end{table*}

\section{Knowledge Encoder}
\label{appendix:knowledge encoder}
% In the paper, we mentioned the use of a Knowledge Encoder module to map propositional formulae into a high-dimensional space. Specifically, during the training steps of the Knowledge Encoder, we also need to convert propositional formulae to d-DNNF format, which will be further explained below:

\textbf{Preliminaries:} This is a brief introduction to d-DNNF, which is used in Knowledge Encoder. A formula that is a conjunction of clauses (a disjunction of literals) is in the Conjunctive Normal Form  (CNF). Let $S$ be the set of propositional variables. A sentence in Negation Normal Form (NNF) is defined as a rooted directed acyclic graph (DAG) where each leaf node is labeled with True, False,  $s, or \neg s, s \in S;$ and each internal node is labeled with $\land$ or $\lor$ and can have discretionarily many children. Deterministic Decomposable Negation Normal Form (d-DNNF) \cite{darwiche2001ddnnf, darwiche2002ddnnf} further imposes that the representation is: (i) \textbf{Deterministic}: An NNF is deterministic if the operands of $\lor$ in all well-formed boolean formula in NNF are mutually inconsistent; (ii) \textbf{Decomposable}: An NNF is decomposable if the operands of $\land$ in all well-formed boolean formula in the NNF are expressed on a mutually disjoint set of variables. Opposite to CNF and more general forms, d-DNNF has many satisfactory tractability properties (e.g., polytime satisfiability and polytime model counting). Because of having tractability properties, it is appealing for complex AI applications to adopt d-DNNF \cite{ddnnfproperties}.

In the paper, we mentioned the use of a \textbf{Knowledge Encoder} module to map propositional formulae into an embedding space. Concretely, we utilize the d-DNNF graph structure to represent a propositional formula $f_{i}$ and then apply a multi-layer Graph Convolutional Network \cite{kipf2017semisupervisedgcn} as an encoder to project the formula, $f_{i}$. In the following paragraphs, we further detail the \textbf{Knowledge Encoder} module. Note that the \textbf{Knowledge Encoder} $\phi_{F}(\cdot)$  is trained before KDAlign framework.

The input for training \( \phi_{F}(\cdot) \) consists of specialized d-DNNF graphs which contribute to enhanced symbolic (knowledge) embeddings. These graphs are built from formulae that have been restructured based on decision paths. To construct the specific graphs based on these formulae, we first change the formulae in CNF and then use \textbf{c2d} to compile these formulae in d-DNNF \cite{darwiche2001ddnnf, darwiche2002ddnnf, c2dcompilerddnnf}. For example, based on Formula `$p_1 \land p_2 \Rightarrow q$' in Section~\ref{sec:knowledge representation}, we construct a CNF expression by Formula.~\eqref{cnfexample}. Then, after executing \textbf{c2d}, Formula.~\eqref{cnfexample} can be expressed in d-DNNF shown by Formula.~\eqref{cnf2ddnnfexample}. 

\begin{equation}
\label{cnfexample}
     \lnot p_1 \lor \lnot p_2 \lor q
\end{equation}

\begin{equation}
\label{cnf2ddnnfexample}
	(\lnot p_1 \land p_2) \lor \lnot p_2 \lor q
\end{equation}

\begin{figure}[h]
\centering
\includegraphics[width=0.25\textwidth]{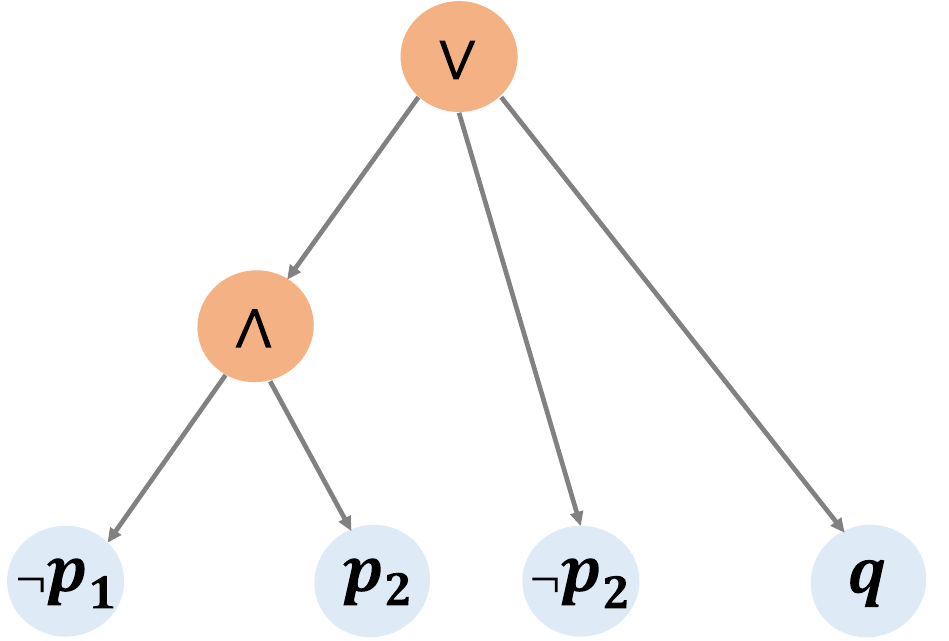}
\caption{The d-DNNF graph structure generated based on Formula.~\eqref{cnf2ddnnfexample}.}
\label{fig:d-DNNF graph structure}
\vspace{-5mm}
\end{figure}

Then a propositional formula can be represented as a directed or undirected graph $G = (V, E)$, consisting of \(N\) nodes denoted by \(v_{i} \in V\) and edges represented as \((v_{i}; v_{j}) \in E\). Individual nodes are either propositions (leaf nodes) or logical operators ($\land$; $\lor$; $\Longrightarrow$), where propositions are connected to their respective operators. Fig.~\ref{fig:d-DNNF graph structure} can help understand the concrete structure. In addition to the mentioned nodes, every graph, like Fig.~\ref{fig:d-DNNF graph structure}, is further augmented by a global node linked to all other nodes. In $\phi_{F}(\cdot)$, the graphs are regarded as undirected graphs.

The layer-wise propagation rule of GCN is,
\vspace{-1mm}
\begin{equation}
    Z^{(l+1)}=\sigma(\tilde{D}^{-\frac{1}{2}}\tilde{A}\tilde{D}^{-\frac{1}{2}}Z^{(l)}W^{(l)})
    \vspace{-1mm}
\end{equation} where $Z^{(l+1)}$ represent the learnt latent node embeddings at $l^{th}$ (note that $Z^{(0)}=X$), $\tilde{A}=A+I_{N}$ represents the adjacency matrix of the undirected graph $G$ with added self-connections through the identity matrix $I_{N}$. $\tilde{D}$ is a diagonal degree matrix with $\tilde{D_{ii}}=\sum_{j}\tilde{A}_{ij}$. The weight matrices for layer-specific training are $W^{(l)}$, and $\sigma(\cdot)$ represents the activation function. To more effectively capture the semantics conveyed through the graphs, the $\phi_{F}(\cdot)$ function incorporates two additional adjustments: heterogeneous node embeddings and semantic regularization, as cited in \cite{xie2019embedding}. The concrete code implementation is accessible at  \url{https://github.com/ZiweiXU/LENSR}.

\begin{table*}[htbp]
\centering
\renewcommand\arraystretch{1.0}
\caption{Optimal Parameter of KDAlign-FeaWAD}
\vspace{-2mm}
\resizebox{0.9\textwidth}{!}{
\begin{tabular}{m{6.75em}<{\raggedleft} | m{2.0em}<{\centering} m{6em}<{\centering} m{2.0em} <{\centering} m{6.5em}<{\centering} m{2.75em}<{\centering} m{5.75em}<{\centering}}
\cmidrule[1.2pt]{1-7}
Dataset Name     & Epoch  & Layers & Learning Rate & Hidden Dimension & Rule Weight & Activation \\
\cmidrule{1-7}
Amazon          &20 &2 &0.001 &32 &0.01 & ReLU  \\
Cardiotocography &20 &2 &0.01 &32 &0.01       & ReLU       \\
Satellite &20 &3 &0.001 &64 &0.05        & ReLU       \\
SpamBase &20 &3 &0.001 &64 &0.05       & ReLU       \\
YelpChi  &100 &2 &0.01 &32 &0.05        & ReLU       \\
\cmidrule[1.2pt]{1-7}
\end{tabular}
}
\label{tab:hyper KDAlign-FeaWAD}
\end{table*}

\begin{table*}[htbp]
\centering
\renewcommand\arraystretch{1.0}
\caption{Optimal Parameter of KDAlign-ResNet}
\vspace{-2mm}
\resizebox{0.9\textwidth}{!}{
\begin{tabular}{m{6.75em}<{\raggedleft} | m{1.5em}<{\centering} m{5.5em}<{\centering} m{2em} <{\centering} m{5em}<{\centering} m{3.0em}<{\centering} m{5.75em}<{\centering} m{3em}<{\centering} m{3em}<{\centering}}
\cmidrule[1.2pt]{1-9}
Dataset Name    & Epoch & Learning Rate & Blocks & Hidden Dimension & Rule Deight & Main Dimension & Dropout First & Dropout Second \\
\cmidrule{1-9}
Amazon           &50 &0.01 &3 &256 &0.1 &192 &0.2 &0              \\
Cardiotocography&50  &0.01 &3 &128 &0.01 &64 &0.2 &0               \\
Satellite       &200 &0.01 &3 &128 &3 &128 &0.2 &0              \\
SpamBase        &50  &0.01 &2 &128 &0.01 &64 &0.2 &0             \\
YelpChi          &50  &0.01 &3 &256 &3 &64 &0.2 &0             \\
\cmidrule[1.2pt]{1-9}
\end{tabular}
}
\label{tab:hyper KDAlign-ResNet}
\end{table*}

\section{Implementation Details}
\label{appendix: implementation}

% \textbf{Implementations of baselines.} All the machine learning methods are implemented by the libraries named \textbf{PyOD}~\cite{zhao2019pyod} and \textbf{scikit-learn}~\cite{pedregosa2011scikit}, and we use the default parameters in ~\cite{han2022adbench,zhao2023admoe}. For deep learning methods, \textbf{ResNet} and \textbf{FTTransformer} are implemented based on their original paper~\cite{gorishniy2021FTTransformerResNet}, and \textbf{MLP} is implemented by the linear layer and ReLU activation function in \textbf{PyTorch}~\cite{paszke2019pytorch}. \textbf{KEnsemble-} methods first compute the anomalous scores of samples by rules (i.e., all-right abnormal paths), where the samples can match the rules, then compute the anomalous scores of these samples by the base method (e.g., XGBOD and MLP), and finally compute the average of the two types of scores as the output. The samples that can not match rules are just detected by the base method. \textbf{KNN-} methods consist of two parts, an unsupervised KNN learner and a deep learning method. KNN learner, which is implemented by \textbf{scikit-learn}~\cite{pedregosa2011scikit} based on the default parameters, helps to find the nearest rule-sample pair and compute the distance of every pair. Since the distance is not differentiable, the distance is just added to the loss function of the deep learning methods as a regularization term for improving the performance.

\textbf{Hardware Specifications}. All our experiments were carried out on a Linux server equipped with AMD EPYC 7763 64-Core Processor, 503GB RAM, and eight NVIDIA RTX4090 GPUs with a total of 192G memory.

\textbf{Hyperparameter Settings}. 
Table~\ref{tab:hyper KDAlign-FeaWAD} and Table~\ref{tab:hyper KDAlign-ResNet} respectively show our optimal hyperparameter settings of KDAlign-FeaWAD and KDAlign-ResNet utilized in our experiments clearly, which are trained on 10 labeled anomaly samples.

\end{document}